\documentclass[a4paper]{article}
\usepackage{etex}
\usepackage[a4paper,top=3cm,left=3cm,right=2cm,bottom=2cm]{geometry}
\usepackage[english]{babel}
\usepackage{amsmath,amssymb,amsthm}
\usepackage{caption}
\usepackage{csquotes}
\usepackage{fancyhdr}
\usepackage{float}
\usepackage{graphics}
\usepackage{graphicx}
\usepackage{hyperref}
\usepackage{lipsum}
\usepackage[titletoc,title]{appendix}
\usepackage{longtable}
\usepackage{soul}
\usepackage{lscape}
\usepackage{multirow}
\usepackage{natbib}
\usepackage[noend]{algorithmic}
\usepackage[numbered]{algorithm}
\usepackage{pst-all}
\usepackage{setspace}
\usepackage{subcaption}
\usepackage{varwidth}
\usepackage{tikz}
\usepackage{tkz-graph}
\usepackage{url}
\usepackage{placeins}
\usetikzlibrary{arrows,backgrounds,positioning,shapes}
\pdfoutput=1
\hypersetup{
  colorlinks=true,
  linkcolor=black,
  citecolor=black,
  urlcolor=black
}

\allowdisplaybreaks

\floatname{algorithm}{Algorithm}

\newcommand{\up}[1]{\raisebox{1.3ex}[0pt]{#1}}

\begin{document}
\sloppy




\title{A heuristic algorithm for a single vehicle  static bike sharing rebalancing problem}

\author{{\bf F{\'a}bio Cruz, Anand Subramanian}\\
Centro de Inform{\'a}tica\\ Universidade Federal da Para{\'i}ba, CEP 58059-900, Jo{\~a}o Pessoa, Brazil.\\
fabiocba@di.ufpb.br, anand@ct.ufpb.br\\
\\
{\bf Bruno P.  Bruck, Manuel Iori}\\
DISMI, University of Modena and Reggio Emilia, 42122 Reggio Emilia, Italy.\\
bruno.p.bruck@gmail.com, manuel.iori@unimore.it
}

\date{}

\maketitle

\vspace{-0.5cm}
\begin{center}
Working Paper, UFPB -- May 2016 \\
\end{center}
\vspace{0.5cm}

\begin{abstract}
{\singlespacing
\noindent 
The static bike rebalancing problem (SBRP) concerns the task of repositioning bikes among stations in self-service bike-sharing systems. This problem can be seen as a variant of the one-commodity pickup and delivery vehicle routing problem, where multiple visits are allowed to be performed at each station, i.e., the demand of a station is allowed to be split.  Moreover, a vehicle may temporarily drop its load at a station, leaving it in excess or, alternatively, collect more bikes from a station (even all of them), thus leaving it in default. Both cases require further visits in order to meet the actual demands of such station. This paper deals with a particular case of the SBRP, in which only a single vehicle is available and the objective is to find a least-cost route that meets the demand of all stations and does not violate the minimum (zero) and maximum (vehicle capacity) load limits along the tour. Therefore, the number of bikes to be collected or delivered at each station should be appropriately determined in order to respect such constraints. We propose an iterated local search (ILS) based heuristic to solve the problem. The ILS algorithm was tested on 980 benchmark instances from the literature and the results obtained are quite competitive when compared to other existing methods. Moreover, our  heuristic was capable of finding most of the known optimal solutions and also of improving the results on a number of open instances.

\vspace{\baselineskip}
}
\end{abstract}

%
%
%
\onehalfspace


\section{Introduction}
\label{sec:Introduction}

The task of repositioning a commodity from one location to another is a well-known problem arising in different contexts such as logistics, transportation, and various disciplines, notably industrial engineering and operations management.
A practical application arises in self-service bike sharing systems (BSS), which are becoming increasingly popular in recent years. Users rent bikes and return them at stations distributed over a region. A vehicle with limited load capacity then periodically collects and delivers bikes across different stations so as to rebalance the system.

Alternatives to the street traffic are important not only because of its impact in urban congestion, but also in the environment, commuting, and so on. The emerging worldwide BSS are proving to be an effective solution to mitigate the effects of traffic issues in large urban centers. Up to 2009, there were about 120 bike sharing programs around the world and, according to \citet{DeMaio:2009}, they have a favorable impact on: decreasing traffic congestion, improving public health, and helping reducing the level of CO$_{2}$ emissions. One of the most famous systems is the \textit{V{\'e}lib'} system in Paris, with 1800 stations and more than 20,000 bikes.

In such systems, each station has an inventory with a load capacity, an initial number of bikes, and consequently a number of free slots where users can return bikes to the system. Throughout the day, some stations may have no bike to be rented or free slots to store returned bikes. Therefore, an attempt to avoid this scenario, which is unpleasant for users, is to determine an initial acceptable number of bikes (and free slots) at each station. This task can be done based on demand history and peaks at each station.

The activity of repositioning bikes among stations on a regular basis is called rebalancing, and this is done by one or more vehicles that move bikes from one station to another in order to restore its inventory to the initial desired configuration. As per \citet{DeMaio:2009}, good rebalancing systems are present in successful bike sharing programs, and since the vehicles move back and forth across an urban area, a vehicle routing optimization can be utilized. 


The rebalancing is either static, performed when nearly no bikes are being used, or dynamic, which is done while the system is still in use. The static bike rebalancing problem (SBRP) is motivated by the fact that very few bikes are being used at night. Indeed, according to \citet{chemla} and \citet{DellAmico2014}, there are many systems that are even closed during the night.

In this work we consider the single vehicle SBRP, which is clearly  $\mathcal{NP}$-hard, because it includes, among others, the classical traveling salesman problem (TSP) as a special case. The SBRP can be seen as a variant of the one-commodity pickup and delivery TSP \citep{HerS04a,HerS04b}, where multiple visits are allowed to be performed at each station, i.e., the demand of a station is allowed to be split.
Moreover, a vehicle may arbitrarily drop its load at a station, leaving it in excess or, alternatively, collect more bikes from a station (even all of them), thus leaving it in default. Both cases require further visits in order to meet the actual demands of such station. This strategy of allowing a station to act as a buffer or a temporary depot is denoted as \emph{preemption} or \emph{temporary operation} (i.e., temporary pickup and temporary dropoff). Finally, visits to balanced stations are optional for the SBRP. \citet{salazar} considered a similar yet different problem, in which an upper limit is imposed on the number of visits to the customers and to the depot, and the single vehicle that performs the rebalancing is not forced to leave the depot with an empty load.




An increasing number of works regarding bike sharing systems and related issues, such as the balancing of their stations, has been published over the last years. Exact approaches for multiple vehicle SBRPs were suggested by \citet{DellAmico2014,di2013constraint,gaspero:2013,di2015balancing,kloimullner2015cluster,raviv:2013}. Moreover, \citet{alvarez2015optimizing,DellAmico2016,forma20153,papazek2014balancing,papazek:2013,raidl:2013,rainer:2013,rainer:2014} also addressed different types of multiple vehicle SBRPs, but with heuristics.

Several exact \cite{benchimol,chemla,erdogan,erdogan:laporte:calvo,salazar} and heuristic \citep{Ho2014180,pal2015free} algorithms were proposed for single vehicle SBRPs. 
Furthermore, in contrast to static rebalancing, there are relatively few works related to dynamic rebalancing 
\cite{caggiani2013dynamic,chemla:hal,contardo2012,kloimullner,schuijbroek:2013}.

The works of \citet{chemla} and \citet{erdogan} were the only ones to consider the same variant dealt in the present paper.  \citet{chemla} proposed a mathematical formulation over an extended graph, where each station is replicated according to an upper bound on the number of visits. Due to its visible intractability, two relaxations were developed. The authors also presented among other contributions, a polynomial algorithm to compute optimal bike displacements for a given sequence of vertices, which is useful to determine if a route is feasible or not, as well as tabu search heuristics and a branch-and-cut algorithm that solves a relaxation of the problem. \citet{erdogan} proposed the first exact method for the problem,  which consists of a branch-and-cut algorithm that makes use of no-good cuts, and they reported optimal solutions for instances with up to 60 stations.


Despite the advances on the development of efficient exact approaches for SBRPs, heuristic methods still appear to be more suitable for dealing with medium and large size instances of this challenging class of problems. In addition, high quality heuristic solutions may be crucial to improve the runtime performance of exact algorithms.

This work proposes a hybrid iterated local search (ILS) based heuristic for the single vehicle SBRP considered in \citet{chemla} and \citet{erdogan}. Hybridized ILS algorithms, especially when combined with randomized variable neighborhood descent (RVND), revealed to be very effective when solving a large variety of vehicle routing problems \cite{DellAmico2016,Pennaetal2013, silva:2015, subramanian2012heuristic, subramanian2010parallel,Vidaletal2015}, including those involving only a single vehicle \citep{Blum03metaheuristics,SubramanianBattarra2013}.


The algorithm that was developed combines successful ingredients from previous works with some problem-specific procedures suggested in \citet{chemla} to improve the solution quality, as well as to check if a solution is infeasible. We also implemented several perturbation mechanisms and the impact of each possible combination on the solution quality and CPU time are measured by extensive computational experiments on a subset of challenging test-problems. The results obtained on 980 benchmark instances from the literature show that our algorithm is quite competitive when compared to other methods, and a number of new best known solutions is reported. We also conduct an analysis on how the performance of the algorithms in terms of solution quality and CPU time varies according to the number of stations and the vehicle capacity.


The remainder of the paper is organized as follows. Section \ref{sec:ProblemDescription} presents a formal problem definition. Section \ref{sec:Heuristic} describes the proposed heuristic algorithm. Section \ref{sec:CompExp} reports and discusses the computational results, and Section \ref{sec:Conclusions} contains the concluding remarks.

\section{Problem description}\label{sec:ProblemDescription}

Let $n$ be the number of stations, $V = \{1, ..., n\}$ be the set of vertices representing their locations (station $0$ represents the depot), and $A$ be the set of arcs in a complete and directed graph $G = (V \cup \{0\}, A)$. For each arc $a_{(i, j)} \in A$, there is a cost $c_{a}$, satisfying the triangular inequality ($c_{(i,j)} + c_{(j,k)} \geq c_{(i,k)}, \forall i, j, k \in V$).

For each $i \in V$, let $p_{i} \in \mathbb{Z}$ be the amount of bikes initially stored,  
$p^{\prime}_{i} \in \mathbb{Z}$ be the number of bikes requested by $i$ after the service is performed, and $d_{i} = p^{\prime}_{i} - p_{i}$ be the total demand. When $d_{i} > 0$ and $d_{i} < 0$, we assume that $i \in V$ is a delivery  and a pickup station, respectively.  
A station $i \in V$ may have no demand ($p_{i} = p'_{i}$) and in this case the visit becomes optional. Each station $i \in V$ has a capacity $q_{i} \in \mathbb{Z}$ and the depot is assumed to have no bikes, i.e., $q_{0} = p_{0} = p^{\prime}_{0} = 0$. Finally, let $Q \in \mathbb{Z}$ be the vehicle capacity.


The objective is to find a least-cost route that starts and ends at the depot, visits each station with non-zero demand at least once, meets the demands of all stations (i.e., the initial load $p_{i}$ is changed to the target demand $p^{\prime}_{i}, \forall i \in V$), and does not violate the minimum (zero) and maximum ($Q$) load limits. Therefore, the number of bikes to be collected or delivered at each visit to a station should be appropriately determined in order to respect such constraints.

Finally, stations may serve to perform temporary operations (preemption), either as a temporary depot or a temporary buffer, i.e., supply more bikes than their initial demand or hold more bikes (without exceeding its inventory load capacity), and in both cases have their demand satisfied in later visits. 

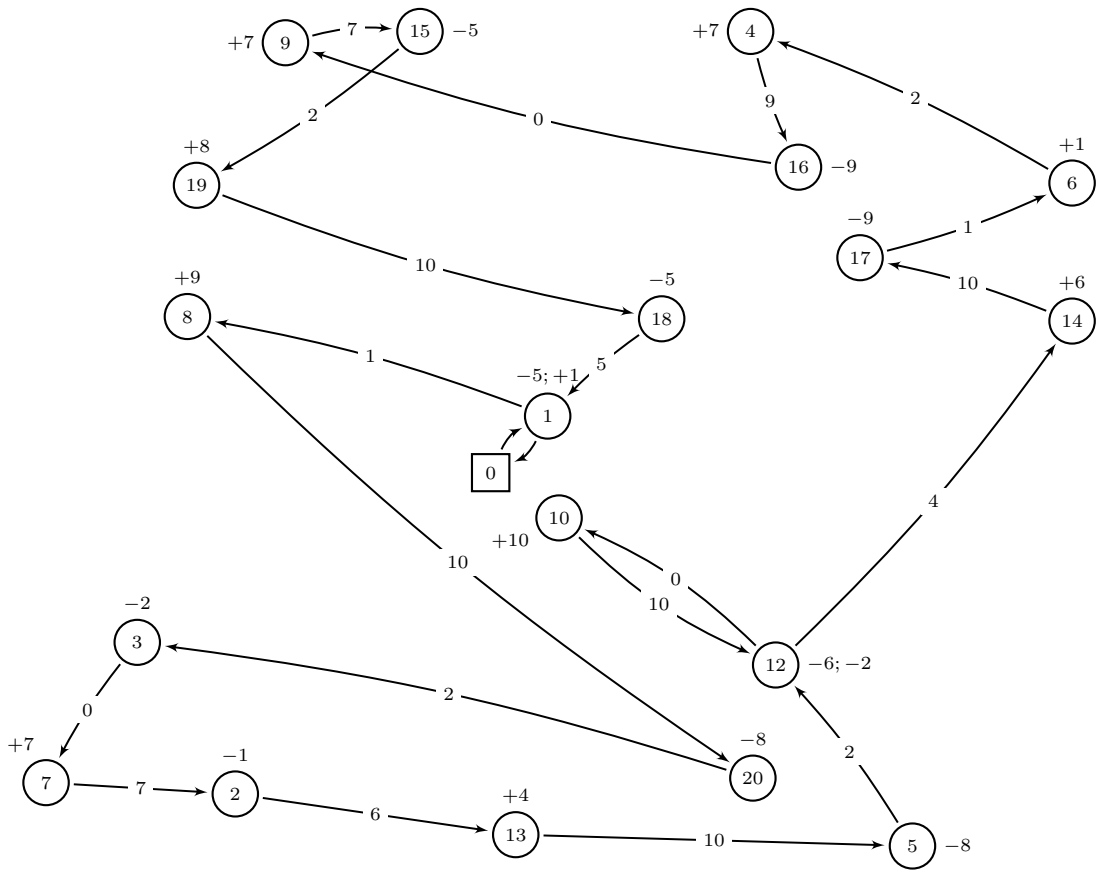
\begin{figure}[!ht]%
\centering%
\begin{tikzpicture}[>=latex',scale=0.3,style={
main node/.style={circle,draw,minimum size=17,inner sep=1pt,thick},
depot node/.style={rectangle,draw,minimum size=14,inner sep=1pt,thick}}]
\SetUpEdge[lw = 0.75pt,color = black,labelcolor = white]
\tikzset{EdgeStyle/.style={->,shorten <=1.5pt,shorten >=1.5pt}}
\scriptsize
\node[depot node](d) at (225mm,225mm) {0};
\node[main node,label={$-5;+1$}](0) at (250mm,250mm){1};
\node[main node,label={$-1$}](1) at (113mm,83mm){2};
\node[main node,label={$-2$}](2) at (70mm,150mm){3};
\node[main node,label={180:$+7$}](3) at (339mm,420mm){4};
\node[main node,label={0:$-8$}](4) at (410mm,60mm){5};
\node[main node,label={$+1$}](5) at (480mm,353mm){6};
\node[main node,label={100:$+7$}](6) at (30mm,88mm){7};
\node[main node,label={$+9$}](7) at (92mm,294mm){8};
\node[main node,label={180:$+7$}](8) at (135mm,415mm){9};
\node[main node,label={200:$+10$}](9) at (255mm,205mm){10};
\node[main node,label={0:$-6;-2$}](11) at (350mm,140mm){12};
\node[main node,label={$+4$}](12) at (236mm,65mm){13};
\node[main node,label={$+6$}](13) at (480mm,292mm){14};
\node[main node,label={0:$-5$}](14) at (194mm,420mm){15};
\node[main node,label={0:$-9$}](15) at (360mm,360mm){16};
\node[main node,label={$-9$}](16) at (387mm,320mm){17};
\node[main node,label={$-5$}](17) at (300mm,293mm){18};
\node[main node,label={$+8$}](18) at (96mm,352mm){19};
\node[main node,label={$-8$}](19) at (340mm,90mm){20};
\Edge[style={bend left=20},label={}](d)(0)
\Edge[style={bend left=20},label={}](0)(d)
\Edge[style={bend right=5},label={$1$}](0)(7)
\Edge[style={bend right=5},label={$10$}](7)(19)
\Edge[style={bend right=5},label={$2$}](19)(2)
\Edge[style={bend right=5},label={$0$}](2)(6)
\Edge[style={bend right=0},label={$7$}](6)(1)
\Edge[style={bend right=0},label={$6$}](1)(12)
\Edge[style={bend right=0},label={$10$}](12)(4)
\Edge[style={bend right=5},label={$2$}](4)(11)
\Edge[style={bend right=10},label={$0$}](11)(9)
\Edge[style={bend right=10},label={$10$}](9)(11)
\Edge[style={bend right=5},label={$4$}](11)(13)
\Edge[style={bend right=5},label={$10$}](13)(16)
\Edge[style={bend right=5},label={$1$}](16)(5)
\Edge[style={bend right=5},label={$2$}](5)(3)
\Edge[style={bend right=5},label={$9$}](3)(15)
\Edge[style={bend left=5},label={$0$}](15)(8)
\Edge[style={bend left=10},label={$7$}](8)(14)
\Edge[style={bend left=5},label={$2$}](14)(18)
\Edge[style={bend right=5},label={$10$}](18)(17)
\Edge[style={bend right=5},label={$5$}](17)(0)
\end{tikzpicture}
%
\caption{Representation of optimal solution with value $5989$ for instance n20q10D}%
\label{figure:sol-represent}%
\end{figure}

Figure \ref{figure:sol-represent} shows a graphical representation of an optimal solution for the benchmark instance n20q10D ($n = 20$ and $Q = 10$). The nodes are distributed according to the spatial coordinates of the stations. The positive and negative values next to the nodes are the number of bikes collected and delivered, respectively. The arcs and their associated values represent the vehicle traveling to the next station in the sequence and the vehicle load, respectively.
For example, the vehicle delivers $2$ bikes in the first visit to station $12$, collects $10$ at station $10$, returns to $12$ to deliver $6$ more (meeting the demand of $8$) and then travels to station $14$ with a load of $4$ bikes. 

\section{Proposed heuristic}
\label{sec:Heuristic}

ILS iteratively alternates between local search (intensification) and perturbation (diversification) mechanisms with a view of finding high quality solutions. In our case, we embed a variable neighborhood descent (VND) \citep{mladenovic1997variable} based procedure in the local search phase of the metaheuristic.
As in previous works (e.g., \citealp{Pennaetal2013}, and \citealp{silva:2015}), the neighborhoods of our algorithm are examined in a random ordering during the search (RVND). 

As opposed to most of the former ILS implementations cited in Section \ref{sec:Introduction}, infeasible solutions are temporary accepted after the application of perturbation moves, not only for the sake of diversification, but also as an attempt to escape from local optimal solutions. This modification was crucial for the favorable performance of the heuristic when dealing with the single vehicle SBRP considered here, which appears to be more challenging to solve than other VRPs where ILS was successfully applied to obtain high quality solutions by only considering the feasible search space. 


\begin{figure}[!ht]
\centering
\begin{subfigure}{.8\textwidth}
\scalebox{0.32}{

\begin{tikzpicture}[scale=18.5,style={
initend node/.style={ellipse, draw, aspect=2, fill=gray!5,text centered, minimum size = 10, inner sep=15pt, thick},
decision node/.style={diamond, draw, aspect=2, fill=gray!5, minimum size=5, text centered, inner sep=15pt,thick},
block node/.style={rectangle, draw, fill=gray!5, minimum size = 10, text centered, inner sep=15pt, thick},
EdgeStyle/.style={->,>=triangle 45,shorten <=1.5pt,shorten >=1.5pt,line width=2pt},
arrow/.style={->,>=triangle 45,shorten <=1.5pt,shorten >=1.5pt,line width=2pt}}]
\Huge
\node[initend node](1) at ( 37pt , 95pt ) [draw, align=center]{ILS$_{\mbox{SBRP}}$};
\node[block node](2) at ( 37pt, 89.5pt ) {$iter_R = 0$};
\node[decision node](3) at ( 37pt,83pt) {$iter_R <  I_R$ ?};
\node[block node](4)  at ( 37 pt,74pt) {$iter_{ILS} = 0$};
\node[initend node](5)  at ( 57 pt,83pt) { Stop };
\node[block node](6)  at ( 37 pt,68.9pt) {Generate initial solution \textbf{$s$}};
\node[decision node](7)  at ( 37 pt,61pt) {$iter_{ILS} < I_{ILS}$ ?};
\node[decision node](8)  at ( 52 pt,53pt) {Is \textbf{$s$} feasible?};
\node[decision node](9)  at ( 21 pt,53pt) {$f($\textbf{$s^\prime$}$) < f$\textbf{$^\star$} ?};
\node[block node](10)  at ( 52 pt,44pt) {\textbf{$s$} = RVND(\textbf{$s$})};
\node[block node](11)  at ( 64 pt,48pt) {\textbf{$s$} = AddUnbalanced(\textbf{$s$})};
\node[decision node](12)  at ( 52 pt,37pt) {$f($\textbf{$s$}$) < f($\textbf{$s^\prime$}$)$ ?};
\node[block node](13)  at ( 52 pt,28pt) {\textbf{$s^\prime$} = \textbf{$s$}};
\node[block node](14)  at ( 52 pt, 18pt) {\textbf{$s$} = Perturb(\textbf{$s^\prime$})};
\node[block node](15)  at ( 52 pt, 23pt) {$iter_{ILS} = 0$};
\node[block node](16)  at ( 37 pt, 13pt) {$iter_{ILS} = iter_{ILS} + 1$};
\node[block node](17)  at ( 21 pt, 43pt) {\textbf{$s^\star$} = \textbf{$s^\prime$}};
\node[block node](18)  at ( 21 pt, 35pt) {$iter_R = iter_R + 1$};
\Edge[style={bend right=0}](1)(2)
\Edge[style={bend right=0}](2)(3)
\Edge[style={bend right=0},label={No}](3)(5)
\Edge[style={bend right=0,pos=0.4},label={Yes}](3)(4.north)
\Edge[style={bend right=0}](4)(6)
\Edge[style={bend right=0}](6)(7)
\Edge[style={bend right=0},label={Yes}](7)(8)
\Edge[style={bend right=0},label={No}](7)(9)
\draw [arrow] (8) -| node[fill=white]{No} (11);
\draw [arrow] (8) -- node[pos=0.4,fill=white]{Yes} (10);
\draw [arrow] (11) |- (10);
\Edge[style={bend left=0}](10)(12)
\Edge[style={pos=0.4},label={Yes}](12)(13)
\draw [arrow] (12) -| node[fill=white]{No} (64pt,20pt) |- (14);
\Edge[style={}](13)(15)
\Edge[style={}](15)(14)
\Edge[style={}](14)(16.north east)
\Edge[style={}](16)(7)
\Edge[style={pos=0.4},label={Yes}](9)(17)		
\draw [arrow] (9) -| node[fill=white]{No} (32pt,43pt) |- (18); 
\Edge[style={}](17)(18)                           
\draw [arrow] (18.west)--(11pt,35pt)|-(3);
\end{tikzpicture}
}
\end{subfigure}
\caption{ILS$_{\text{SBRP}}$ flowchart}
\label{figure:ils-flowchart}
\end{figure}
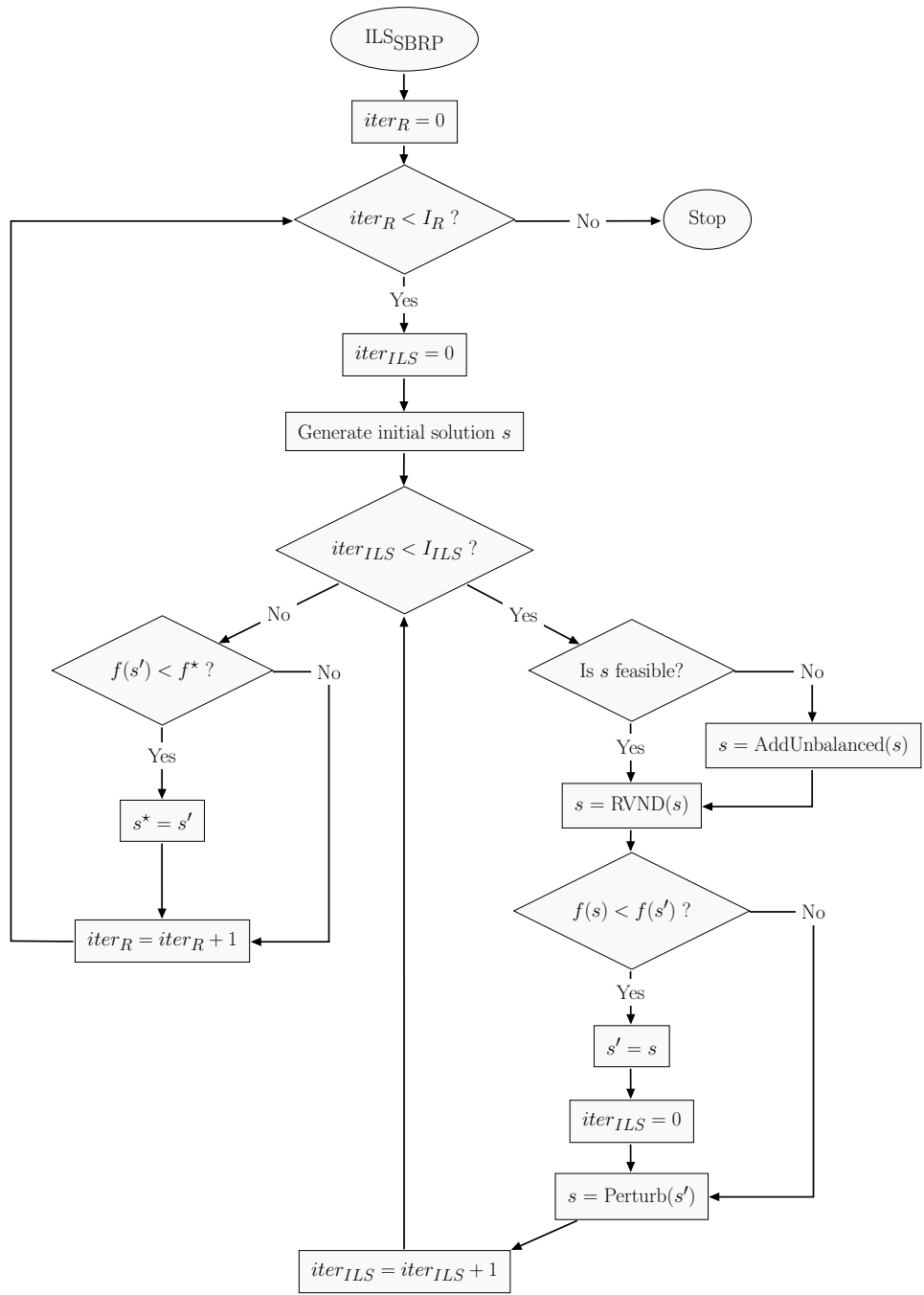

The proposed hybrid heuristic, called ILS$_{\text{SBRP}}$, combines multiple restarts, local search, perturbations mechanisms, and a repair phase. Figure \ref{figure:ils-flowchart} illustrates the flowchart of ILS$_{\text{SBRP}}$. For each of the $I_{R}$ restarts, a feasible initial solution is generated using a simple greedy randomized constructive algorithm (see Section \ref{sec:Construction}). Next, local search, perturbation and repair procedures are successively applied until the stopping criterion is met, that is, when the number of consecutive attempts to escape from a local optimal solution reaches $I_{ILS}$ trials. Because perturbation moves are allowed to produce infeasible solutions, we implemented a procedure called AddUnbalanced \citep{chemla} (see Section \ref{sec:InfeasSol}) with the aim of repairing such solutions. Nevertheless, there is no guarantee that a solution will be feasible after applying this procedure. When an infeasible solution is not totally repaired and the local search does not find a move that leads to a feasible solution, then the infeasible solution is disregarded. Note that perturbation is always applied over the best solution of the current multi-start iteration. Finally, ILS$_{\text{SBRP}}$ returns the best solution found among all restarts.


\subsection{Solution representation}\label{sec:SolRepresent}

A solution for the single vehicle SBRP considered in the present work can be represented as a sequence of visits to stations, starting and ending at the depot, along with the amount of bikes collected or delivered at each visit.

Three vectors are used as data structures to store: (i) the route, where the first and last element are fixed at $0$, i.e., the depot; (ii) the operation performed by the vehicle during a visit, where negative and positive values indicate the amount of bikes delivered and collected, respectively; and (iii) the vehicle load during the route.

As in \citet{chemla}, a flow network is used to check in polynomial time whether or not a solution is feasible, with respect to bike displacements and vehicle capacity, given a sequence of vertices representing visits to stations. A detailed explanation can be found in Appendix \ref{app:checking}.

We also use another data structure which consists of a key-value map composed by $n+1$ elements that store the number of visits performed at each station. This is useful, for example, to check whether a solution includes all stations with non-zero demand. Note that information hold in (ii) is extracted from the computed bike displacements when solving the max-flow problem (see Appendix \ref{app:checking}). From such, it is possible to derive, in linear time, the vehicle loads in (iii) by the adding or subtracting the bike displacements at each visit.


\subsection{Repairing infeasible solutions}\label{sec:InfeasSol}

As already mentioned, infeasible solutions are allowed after perturbations. We therefore implemented  a procedure called AddUnbalanced \citep{chemla}, which tries to repair a solution by adding stations to the route. More precisely, both the most unbalanced station in excess and in default, $i$ and $j$, respectively, are selected and three moves are possible: (i) adding arcs $(j,i)$ and $(i,j)$ after the existing visit to $j$; (ii) adding arcs $(i,j)$ and $(j,i)$ after the existing visit to $i$; (iii) if both $i$ and $j$ are not in the sequence, adding $(i,j)$ at the end of the sequence, before returning to the depot.

For example, let us consider a scenario where stations $i = 12$ and $j = 14$ are the most unbalanced. More precisely $i$ has initially $20$ bikes and a demand of $-10$, i.e., a pickup station, while $j$ is initially holding $3$ out of $10$ (target) bikes, i.e., a delivery station with demand $7$. An infeasible solution is presented in Figure \ref{figure:infeas-sol}, where the referred stations are not balanced, that is, their demands are not met, since only $4$ bikes were collected in station $12$ and 4 bikes were delivered at station $14$. Figure \ref{figure:addunb} shows a modified solution, where after the addition of arcs $(14,12)$ and $(12,14)$, a new and feasible configuration of bike displacements were determined by means of the maximum flow check (see Appendix \ref{app:checking}). We can observe that the second visit complements the first one, meeting the demand of both stations: the vehicle deliveries $1$ bike at station $14$, collects the remaining $7$ at station $12$, now balanced, and finally meets the demand of station $14$ by delivering $6$ more bikes.

{\centering
\begin{figure}[!ht]
\centering
\begin{subfigure}{1.\textwidth}
    \centering
    \begin{tikzpicture}[>=latex',scale=0.7,style={
main node/.style={circle,draw,minimum size=17,inner sep=1pt,thick},
depot node/.style={rectangle,draw,minimum size=10,inner sep=1pt,thick},
cloud node/.style={cloud,fill=gray!0,draw,cloud puffs=10,cloud puff arc=120,aspect=2,inner ysep=1em}}]
\SetUpEdge[lw = 0.75pt,color = black,labelcolor = white]
\tikzset{EdgeStyle/.style={->,shorten <=1.5pt,shorten >=1.5pt}}
\scriptsize
\node[main node,label={$+8$}](2) at (7,2.26){3};
\node[main node,label={$+8$}](5) at (5.52,0.7){6};
\node[cloud node,label={}](6) at (0,6){...};
\node[main node,label={360:$-2$}](10) at (1.5,4){11};
\node[main node,label={270:$+4$}](11) at (0,0.5){12};
\node[main node,label={180:$-4$}](13) at (13.1,2.6){14};
\node[main node,label={90:$-5$}](14) at (14.76,0.18){15};
\node[cloud node,label={}](17) at (14,6){...};
\node[main node,label={360:$-9$}](19) at (10,4.7){20};
\Edge[style={bend right=5},label={$0$}](6)(11)
\Edge[style={bend right=5},label={$4$}](11)(10)
\Edge[style={bend right=10},label={$2$}](10)(5)
\Edge[style={bend right=20,pos=0.20},label={$10$}](5)(19)
\Edge[style={bend right=5},label={$1$}](19)(2)
\Edge[style={bend right=5},label={$9$}](2)(14)
\Edge[style={bend right=5},label={$4$}](14)(13)
\Edge[style={bend right=5},label={$0$}](13)(17)
\end{tikzpicture}
    \caption{Infeasible solution}
    \label{figure:infeas-sol}
\end{subfigure}
\begin{subfigure}{1.\textwidth}
    \begin{tikzpicture}[>=latex',scale=0.7,style={
main node/.style={circle,draw,minimum size=17,inner sep=1pt,thick},
depot node/.style={rectangle,draw,minimum size=10,inner sep=1pt,thick},
cloud node/.style={cloud,fill=gray!0,draw,cloud puffs=10,cloud puff arc=120,aspect=2,inner ysep=1em}}]
\SetUpEdge[lw = 0.75pt,color = black,labelcolor = white]
\tikzset{EdgeStyle/.style={->,shorten <=1.5pt,shorten >=1.5pt}}
\scriptsize
\node[main node,label={180:$+8$}](2) at (7,2.26){3};
\node[main node,label={$+8$}](5) at (5.52,0.7){6};
\node[cloud node,label={}](6) at (0,6){...};
\node[main node,label={360:$-2$}](10) at (1.5,4){11};
\node[main node,label={270:$+3;+7$}](11) at (0,0.5){12};
\node[main node,label={360:$-1;-6$}](13) at (13.1,2.6){14};
\node[main node,label={90:$-5$}](14) at (14.76,0.18){15};
\node[cloud node,label={}](17) at (14,6){...};
\node[main node,label={$-9$}](19) at (10,4.7){20};
\Edge[style={bend right=5},label={$0$}](6)(11)
\Edge[style={bend right=0},label={$3$}](11)(10)
\Edge[style={bend right=5,pos=0.6},label={$1$}](10)(5)
\Edge[style={bend right=20,pos=0.2},label={$9$}](5)(19)
\Edge[style={bend right=20,pos=0.3},label={$0$}](19)(2)
\Edge[style={bend right=5,pos=0.6},label={$8$}](2)(14)
\Edge[style={bend right=5},label={$3$}](14)(13)
\Edge[style={bend right=22},label={$2$}](13)(11)
\Edge[style={bend right=18},label={$9$}](11)(13)
\Edge[style={bend right=5},label={$3$}](13)(17)
\end{tikzpicture}
    \centering
    \caption{Feasible solution after additional visits to unbalanced stations $12$ and $14$}
    \label{figure:addunb}
\end{subfigure}
\caption{Handling an infeasible solution by considering additional visits to unbalanced stations}
\end{figure}
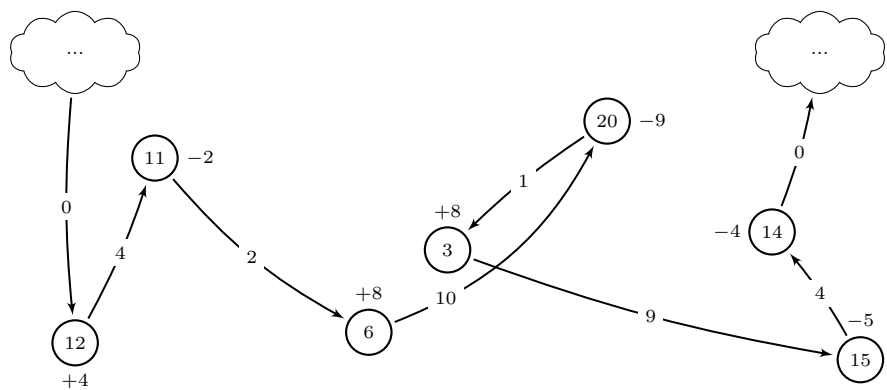
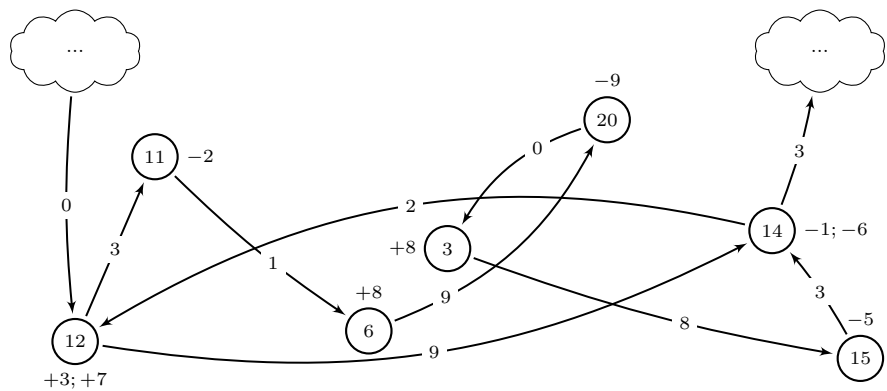}

It is worth emphasizing that the AddUnbalanced procedure does not necessary lead to a feasible solution. However, in general, experimental results showed that such procedure has a high level of success in fully repairing infeasible solutions.




\subsection{Constructive Procedure}\label{sec:Construction}

The pseudocode of the greedy randomized constructive procedure is presented in Alg. \ref{alg:initialSol}. The algorithm stores and maintains a list of open vertices ($OV$) corresponding to stations whose demands are still not fully met. Stations without demand are also included in this list.  In order to ensure a level of diversity during the process of generating an initial solution, $OV$ is sorted in random order (line \ref{alg:initialSol:OV}).

The algorithm follows a greedy procedure by selecting the first vertex to be inserted at the end of the route (before the depot) whose demand is completely met by a single visit without violating the load limits ($[0,Q]$). 
Next, the vehicle load is updated and the station that was inserted into the partial solution is removed from $OV$ (lines \ref{alg:initialSol:if1}-\ref{alg:initialSol:endIf1}).

However, it may come to a point where no station can be fully served in a single visit, either because the vehicle has not enough bikes to deliver, or the residual capacity is not sufficient to  collect the required bikes at once. Hence, a split becomes necessary.  The second part of the constructive procedure (lines \ref{alg:initialSol:if2}-\ref{alg:initialSol:endIf2}) iterates over $OV$ searching for a station whose demand maximizes the number of bikes that can be delivered or collected. Ties are broken according to the nearest insertion criterion. The station demand and vehicle load are updated after the insertion. 
Next, the algorithm restarts from line \ref{alg:initialSol:repeat} and the entire insertion procedure is repeated until  $OV$ becomes empty. Note that the generated initial solution is always feasible.

{\scriptsize
    \begin{algorithm}[!ht]
	\caption{Initial Solution Constructive Procedure}\label{alg:initialSol}
	\begin{algorithmic}[1]
	    \STATE Procedure GenerateInitialSolution
	    \STATE $Q^{\prime}$ $\leftarrow$ $Q$
	    \STATE $Solution \leftarrow \varnothing$
	    \STATE $OV$ $\leftarrow$ List sorted in random order with all stations where $d_i \neq 0$  + random ones with $d_i = 0$\label{alg:initialSol:OV} 
	    \REPEAT \label{alg:initialSol:repeat} 
	    \STATE $inserted$ $\leftarrow$ $false$
	    \FORALL {$i \in OV$} 
	    \IF {$d_{i} \leq Q^{\prime} \; \OR \; Q - Q^{\prime} \geq d_{i}$} \label{alg:initialSol:if1} 
	    \STATE $Solution \leftarrow Solution \cup i$ \label{alg:initialSol:addStation} 
	    \STATE Update vehicle capacity and remove $i$ from $OV$
	    \STATE $inserted$ $\leftarrow$ $true$
	    \STATE $break$
	    \ENDIF \label{alg:initialSol:endIf1} 
	    \ENDFOR
	    \IF {$\NOT \; inserted$} \label{alg:initialSol:if2} 
	    \FORALL {$j \in OV$}
	    \STATE compute $exchange_{j}$
	    \ENDFOR
	    \STATE $i \leftarrow \max \{exchange_{j} \; | \; j \in exchange\}$
	    \STATE $Solution \leftarrow Solution \cup i$
	    \ENDIF \label{alg:initialSol:endIf2} 
	    \STATE update $OV$
	    \STATE update $Q^{\prime}$
	    \UNTIL {$OV \neq \varnothing$}
	    \STATE \textbf{return} $Solution$
	    \STATE \textbf{end} GenerateInitialSolution.
	\end{algorithmic}
    \end{algorithm}
}

\subsection{Local search}\label{sec:LocalSearch}

Initial and perturbed solutions are possibly improved by means of an RVND based procedure during the local search. RVND consists of systematically examining different types of neighborhoods in a random ordering. In particular, if the best neighbor consists of an improving move, then the search may continue from any of the existing neighborhoods (including the last one to be explored) at random. Otherwise, a different neighborhood other than those that did not succeed in finding an improved move is randomly selected. The procedure ends when all neighborhoods fail to refine the current solution. Only feasible moves are accepted.

The following six neighborhood structures were implemented.

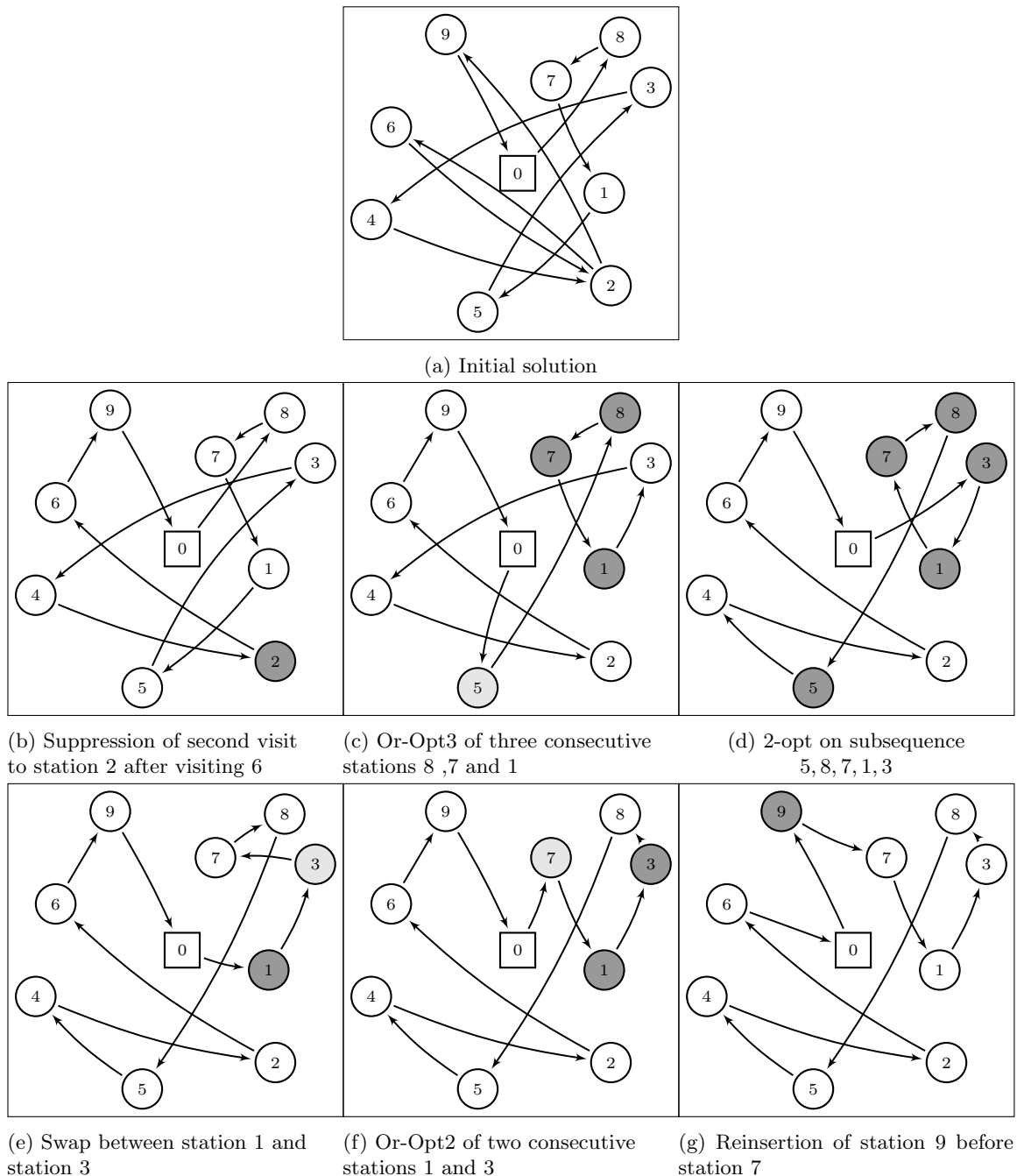
\begin{figure}[!ht]%
\centering
\begin{subfigure}{.315\textwidth}%
\begin{tikzpicture}[>=latex',framed,scale=2,style={
main node/.style={circle,draw,minimum size=17,inner sep=1pt,thick},
depot node/.style={rectangle,draw,minimum size=15,inner sep=1pt,thick}}]
\SetUpEdge[lw = 0.75pt,color = black,labelcolor = white]
\tikzset{EdgeStyle/.style={->,shorten <=1.5pt,shorten >=1.5pt}}
\scriptsize
\node[depot node](0) at (11.5mm,11.5mm){0};
\node[main node](1) at (18.0mm,10.0mm){1};
\node[main node](2) at (18.5mm,3.0mm){2};
\node[main node](3) at (21.5mm,18.0mm){3};
\node[main node](4) at (0.5mm,8.0mm){4};
\node[main node](5) at (8.5mm,1.0mm){5};
\node[main node](6) at (2.0mm,15.0mm){6};
\node[main node](7) at (14.0mm,18.5mm){7};
\node[main node](8) at (19.2mm,21.8mm){8};
\node[main node](9) at (6.1mm,22.0mm){9};
\Edge[style={bend right=7}](0)(8)
\Edge[style={bend right=7}](8)(7)
\Edge[style={bend right=7}](7)(1)
\Edge[style={bend left=10}](1)(5)
\Edge[style={bend left=10}](5)(3)
\Edge[style={bend right=14}](3)(4)
\Edge[style={bend right=7}](4)(2)
\Edge[style={bend right=7}](2)(6)
\Edge[style={bend right=7}](6)(2)
\Edge[style={bend right=10}](2)(9)
\Edge[style={bend left=3}](9)(0)
\end{tikzpicture}%
\caption{Initial solution}%
\label{figure:initial-sol}%
\end{subfigure}

\begin{subfigure}{.315\textwidth}%
\begin{tikzpicture}[baseline=(0.base),framed,>=latex',scale=2,style={
main node/.style={circle,draw,minimum size=17,inner sep=1pt,thick},
depot node/.style={rectangle,draw,minimum size=15,inner sep=1pt,thick},
feat1 node/.style={circle,fill=gray!80,draw,minimum size=17,inner sep=1pt,thick},
feat2 node/.style={circle,fill=gray!20,draw,minimum size=17,inner sep=1pt,thick}}]
\SetUpEdge[lw = 0.75pt,color = black,labelcolor = white]
\tikzset{EdgeStyle/.style={->,shorten <=1.5pt,shorten >=1.5pt}}
\scriptsize
\node[depot node](0) at (11.5mm,11.5mm){0};
\node[main node](1) at (18.0mm,10.0mm){1};
\node[feat1 node](2) at (18.5mm,3.0mm){2};
\node[main node](3) at (21.5mm,18.0mm){3};
\node[main node](4) at (0.5mm,8.0mm){4};
\node[main node](5) at (8.5mm,1.0mm){5};
\node[main node](6) at (2.0mm,15.0mm){6};
\node[main node](7) at (14.0mm,18.5mm){7};
\node[main node](8) at (19.2mm,21.8mm){8};
\node[main node](9) at (6.1mm,22.0mm){9};
\Edge[style={bend right=0}](0)(8)
\Edge[style={bend right=7}](8)(7)
\Edge[style={bend right=0}](7)(1)
\Edge[style={bend left=7}](1)(5)
\Edge[style={bend left=13}](5)(3)
\Edge[style={bend right=14}](3)(4)
\Edge[style={bend right=7}](4)(2)
\Edge[style={bend left=7}](2)(6)
\Edge[style={bend right=0}](6)(9)
\Edge[style={bend left=3}](9)(0)
\end{tikzpicture}%
\caption{Suppression of second visit \\to station $2$ after visiting $6$}%
\label{figure:sup}%
\end{subfigure}%
\begin{subfigure}{.315\textwidth}%
\begin{tikzpicture}[baseline=(0.base),framed,>=latex',scale=2,style={
main node/.style={circle,draw,minimum size=17,inner sep=1pt,thick},
depot node/.style={rectangle,draw,minimum size=15,inner sep=1pt,thick},
feat1 node/.style={circle,fill=gray!80,draw,minimum size=17,inner sep=1pt,thick},
feat2 node/.style={circle,fill=gray!20,draw,minimum size=17,inner sep=1pt,thick}}]
\SetUpEdge[lw = 0.75pt,color = black,labelcolor = white]
\tikzset{EdgeStyle/.style={->,shorten <=1.5pt,shorten >=1.5pt}}
\scriptsize
\node[depot node](0) at (11.5mm,11.5mm){0};
\node[feat1 node](1) at (18.0mm,10.0mm){1};
\node[main node](2) at (18.5mm,3.0mm){2};
\node[main node](3) at (21.5mm,18.0mm){3};
\node[main node](4) at (0.5mm,8.0mm){4};
\node[feat2 node](5) at (8.5mm,1.0mm){5};
\node[main node](6) at (2.0mm,15.0mm){6};
\node[feat1 node](7) at (14.0mm,18.5mm){7};
\node[feat1 node](8) at (19.2mm,21.8mm){8};
\node[main node](9) at (6.1mm,22.0mm){9};
\Edge[style={bend right=7}](0)(5)
\Edge[style={bend right=7}](5)(8)
\Edge[style={bend right=7}](8)(7)
\Edge[style={bend right=7}](7)(1)
\Edge[style={bend right=7}](1)(3)
\Edge[style={bend right=14}](3)(4)
\Edge[style={bend right=7}](4)(2)
\Edge[style={bend left=7}](2)(6)
\Edge[style={bend right=0}](6)(9)
\Edge[style={bend left=3}](9)(0)
\end{tikzpicture}%
\caption{Or-Opt3 of three consecutive \\ stations $8$ ,$7$ and $1$}
\label{figure:or-opt3}%
\end{subfigure}%
\begin{subfigure}{.315\textwidth}%
\begin{tikzpicture}[baseline=(0.base),framed,>=latex',scale=2,style={
main node/.style={circle,draw,minimum size=17,inner sep=1pt,thick},
depot node/.style={rectangle,draw,minimum size=15,inner sep=1pt,thick},
feat1 node/.style={circle,fill=gray!80,draw,minimum size=17,inner sep=1pt,thick},
feat2 node/.style={circle,fill=gray!20,draw,minimum size=17,inner sep=1pt,thick}}]
\SetUpEdge[lw = 0.75pt,color = black,labelcolor = white]
\tikzset{EdgeStyle/.style={->,shorten <=1.5pt,shorten >=1.5pt}}
\scriptsize
\node[depot node](0) at (11.5mm,11.5mm){0};
\node[feat1 node](1) at (18.0mm,10.0mm){1};
\node[main node](2) at (18.5mm,3.0mm){2};
\node[feat1 node](3) at (21.5mm,18.0mm){3};
\node[main node](4) at (0.5mm,8.0mm){4};
\node[feat1 node](5) at (8.5mm,1.0mm){5};
\node[main node](6) at (2.0mm,15.0mm){6};
\node[feat1 node](7) at (14.0mm,18.5mm){7};
\node[feat1 node](8) at (19.2mm,21.8mm){8};
\node[main node](9) at (6.1mm,22.0mm){9};
\Edge[style={bend right=7}](0)(3)
\Edge[style={bend left=7}](3)(1)
\Edge[style={bend left=7}](1)(7)
\Edge[style={bend left=7}](7)(8)
\Edge[style={bend left=7}](8)(5)
\Edge[style={bend left=7}](5)(4)
\Edge[style={bend right=7}](4)(2)
\Edge[style={bend left=7}](2)(6)
\Edge[style={bend right=0}](6)(9)
\Edge[style={bend left=3}](9)(0)
\end{tikzpicture}%
\caption{2-opt on subsequence \\$5,8,7,1,3$}%
\label{figure:2-opt}%
\end{subfigure}%

\begin{subfigure}{.315\textwidth}%
\begin{tikzpicture}[baseline=(0.base),framed,>=latex',scale=2,style={
main node/.style={circle,draw,minimum size=17,inner sep=1pt,thick},
depot node/.style={rectangle,draw,minimum size=15,inner sep=1pt,thick},
feat1 node/.style={circle,fill=gray!80,draw,minimum size=17,inner sep=1pt,thick},
feat2 node/.style={circle,fill=gray!20,draw,minimum size=17,inner sep=1pt,thick}}]
\SetUpEdge[lw = 0.75pt,color = black,labelcolor = white]
\tikzset{EdgeStyle/.style={->,shorten <=1.5pt,shorten >=1.5pt}}
\scriptsize
\node[depot node](0) at (11.5mm,11.5mm){0};
\node[feat1 node](1) at (18.0mm,10.0mm){1};
\node[main node](2) at (18.5mm,3.0mm){2};
\node[feat2 node](3) at (21.5mm,18.0mm){3};
\node[main node](4) at (0.5mm,8.0mm){4};
\node[main node](5) at (8.5mm,1.0mm){5};
\node[main node](6) at (2.0mm,15.0mm){6};
\node[main node](7) at (14.0mm,18.5mm){7};
\node[main node](8) at (19.2mm,21.8mm){8};
\node[main node](9) at (6.1mm,22.0mm){9};
\Edge[style={bend right=7}](0)(1)
\Edge[style={bend right=7}](1)(3)
\Edge[style={bend right=7}](3)(7)
\Edge[style={bend left=7}](7)(8)
\Edge[style={bend left=7}](8)(5)
\Edge[style={bend left=7}](5)(4)
\Edge[style={bend right=7}](4)(2)
\Edge[style={bend left=7}](2)(6)
\Edge[style={bend right=0}](6)(9)
\Edge[style={bend left=3}](9)(0)
\end{tikzpicture}%
\caption{Swap between station $1$ and\\ station $3$}%
\label{figure:swap}%
\end{subfigure}%
\begin{subfigure}{.315\textwidth}%
\begin{tikzpicture}[baseline=(0.base),framed,>=latex',scale=2,style={
main node/.style={circle,draw,minimum size=17,inner sep=1pt,thick},
depot node/.style={rectangle,draw,minimum size=15,inner sep=1pt,thick},
feat1 node/.style={circle,fill=gray!80,draw,minimum size=17,inner sep=1pt,thick},
feat2 node/.style={circle,fill=gray!20,draw,minimum size=17,inner sep=1pt,thick}}]
\SetUpEdge[lw = 0.75pt,color = black,labelcolor = white]
\tikzset{EdgeStyle/.style={->,shorten <=1.5pt,shorten >=1.5pt}}
\scriptsize
\node[depot node](0) at (11.5mm,11.5mm){0};
\node[feat1 node](1) at (18.0mm,10.0mm){1};
\node[main node](2) at (18.5mm,3.0mm){2};
\node[feat1 node](3) at (21.5mm,18.0mm){3};
\node[main node](4) at (0.5mm,8.0mm){4};
\node[main node](5) at (8.5mm,1.0mm){5};
\node[main node](6) at (2.0mm,15.0mm){6};
\node[feat2 node](7) at (14.0mm,18.5mm){7};
\node[main node](8) at (19.2mm,21.8mm){8};
\node[main node](9) at (6.1mm,22.0mm){9};
\Edge[style={bend right=7}](0)(7)
\Edge[style={bend right=7}](7)(1)
\Edge[style={bend right=7}](1)(3)
\Edge[style={bend right=7}](3)(8)
\Edge[style={bend left=7}](8)(5)
\Edge[style={bend left=7}](5)(4)
\Edge[style={bend right=7}](4)(2)
\Edge[style={bend left=7}](2)(6)
\Edge[style={bend right=0}](6)(9)
\Edge[style={bend left=3}](9)(0)
\end{tikzpicture}%
\caption{Or-Opt2 of two consecutive \\ stations $1$ and $3$} 
\label{figure:or-opt2}%
\end{subfigure}%
\begin{subfigure}{.315\textwidth}%
\begin{tikzpicture}[baseline=(0.base),framed,>=latex',scale=2,style={
main node/.style={circle,draw,minimum size=17,inner sep=1pt,thick},
depot node/.style={rectangle,draw,minimum size=15,inner sep=1pt,thick},
feat1 node/.style={circle,fill=gray!80,draw,minimum size=17,inner sep=1pt,thick},
feat2 node/.style={circle,fill=gray!20,draw,minimum size=17,inner sep=1pt,thick}}]
\SetUpEdge[lw = 0.75pt,color = black,labelcolor = white]
\tikzset{EdgeStyle/.style={->,shorten <=1.5pt,shorten >=1.5pt}}
\scriptsize
\node[depot node](0) at (11.5mm,11.5mm){0};
\node[main node](1) at (18.0mm,10.0mm){1};
\node[main node](2) at (18.5mm,3.0mm){2};
\node[main node](3) at (21.5mm,18.0mm){3};
\node[main node](4) at (0.5mm,8.0mm){4};
\node[main node](5) at (8.5mm,1.0mm){5};
\node[main node](6) at (2.0mm,15.0mm){6};
\node[main node](7) at (14.0mm,18.5mm){7};
\node[main node](8) at (19.2mm,21.8mm){8};
\node[feat1 node](9) at (6.1mm,22.0mm){9};
\Edge[style={bend right=3}](0)(9)
\Edge[style={bend right=7}](7)(1)
\Edge[style={bend right=7}](1)(3)
\Edge[style={bend right=12}](3)(8)
\Edge[style={bend left=7}](8)(5)
\Edge[style={bend left=7}](5)(4)
\Edge[style={bend right=7}](4)(2)
\Edge[style={bend left=7}](2)(6)
\Edge[style={bend right=0}](6)(0)
\Edge[style={bend right=7}](9)(7)
\end{tikzpicture}%
\caption{Reinsertion of station $9$ before station $7$}%
\label{figure:or-opt1}%
\end{subfigure}%
\caption{Example regarding the application of neighborhood structures}%
\end{figure}

\begin{itemize}
    \itemsep.0pt
    \item Reinsertion --- $N^{(1)}$: A station is removed and then reinserted in another position of the sequence.
    \item Or-opt2 --- $N^{(2)}$: Two consecutive stations are removed and then inserted in another position.
    \item Or-opt3 --- $N^{(3)}$: Three consecutive stations are removed and then inserted in another position.
    \item 2-opt --- $N^{(4)}$: Two non-adjacent arcs are removed from the sequence and then two new ones are inserted. In other words, a subsequence of the tour is reversed. 
    \item Swap --- $N^{(5)}$: Permutation of two stations.
    \item Suppression --- $N^{(6)}$: Given a sequence $L = i_{0},i_{1},...,i_{k}$,  a suppression list is composed of visits to stations $i_{j}, \forall j \in \{1,\dots,k-1\}$, such that $p'_{i_{j}} = p_{i_{j}}$ (zero demand) or $p'_{i_{j}} \neq p_{i_{j}}$ and ${i_{j}}$ is visited more than once in the tour.
	The best move, if any, consists in selecting one station to be removed from $L$ so that the solution cost is minimized and the resulting new sequence $L'$ is feasible. For example, Figure \ref{figure:sup} shows the removal of an additional visit to station $2$, thus modifying the subsequence $2,6,2,9,0$ to $2,6,9,0$. This neighborhood was originally proposed by \citet{chemla}, but the authors considered all stations. 
\end{itemize}
        

The first five are well-known TSP neighborhood structures, while the last is a problem-specific neighborhood.  Figure \ref{figure:initial-sol} depicts an initial solution and Figures \ref{figure:sup} to \ref{figure:or-opt1} illustrate modified solutions that were obtained after changing the previous one by means of one of the neighborhoods described above. For example, Figure \ref{figure:2-opt} shows a solution in which a 2-opt move was applied over the solution shown in Figure \ref{figure:or-opt3}. For ease of presentation, 
values of pickup/delivery operations as well as the vehicle load are omitted.

\subsection{Perturbation mechanisms}\label{sec:Perturb}

One of the four mechanisms described below is selected at random whenever the algorithm enters the perturbation phase.

\begin{itemize} 
    \item AddBuffer --- $P^{(1)}$: An additional visit to a station is included,  expecting to act as buffer, using the cheapest insertion criterion. Unrouted stations are inserted twice using the same criterion \citep{chemla}.
    \item AddStations --- $P^{(2)}$: This perturbation mechanism generalizes the previous one in the sense of allowing multiple visits to be added in the solution, but with a different insertion criterion. More precisely, an additional visit (or two, in the case of unrouted stations) to up to three random stations are included towards the end of the route. Here we only consider stations that are visited at most once. Adjacent visits to the same station are forbidden.
    \item Double-Bridge --- $P^{(3)}$: Introduced by \citet{Martin91} for the TSP, this perturbation consists of a permutation of two subsequences. As a result, four arcs are removed and four new ones are added so as to generate a new sequence.        
    \item Suppression --- $P^{(4)}$: A suppression move (see Section \ref{sec:LocalSearch}) is applied at random, but in this case the resulting modified sequence is allowed to be infeasible.
\end{itemize}

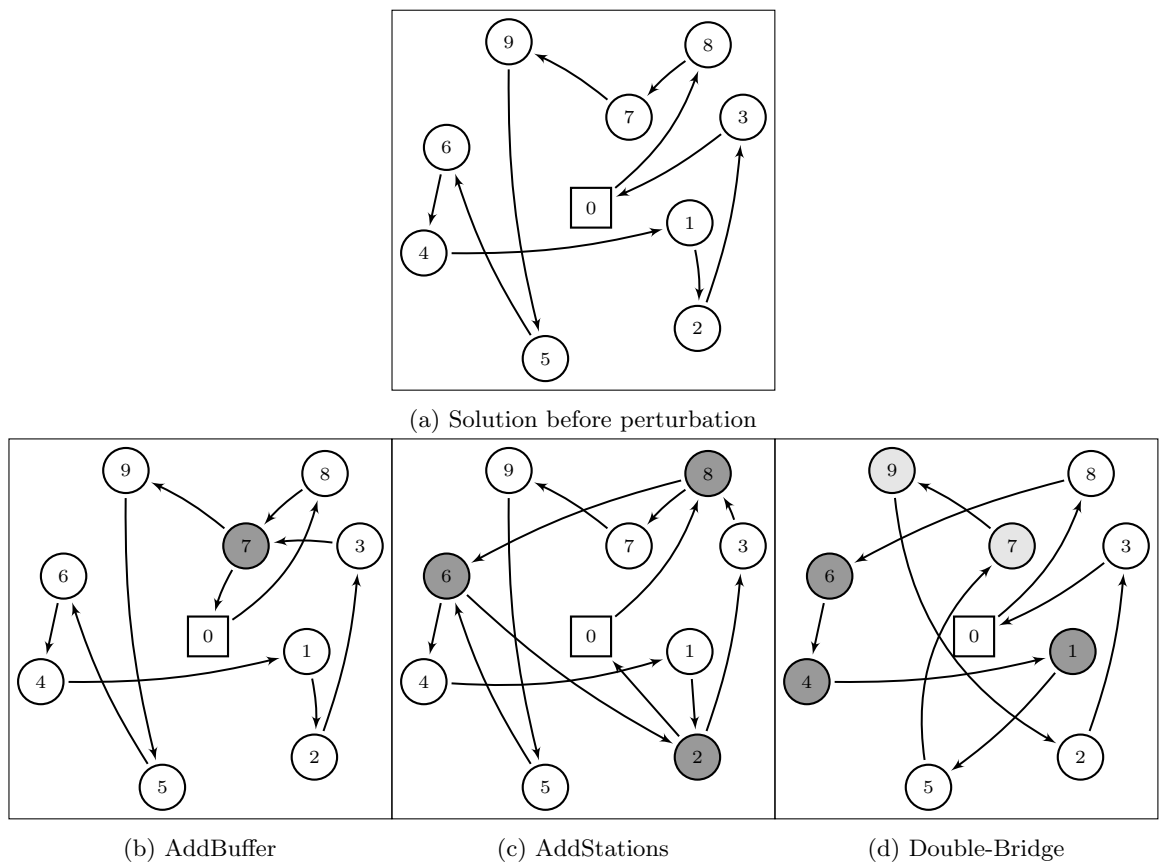
\begin{figure}[!ht]
\centering
\begin{subfigure}{.315\textwidth}%
\begin{tikzpicture}[>=latex',framed,scale=2,style={
main node/.style={circle,draw,minimum size=17,inner sep=1pt,thick},
depot node/.style={rectangle,draw,minimum size=15,inner sep=1pt,thick}}]
\SetUpEdge[lw = 0.75pt,color = black,labelcolor = white]
\tikzset{EdgeStyle/.style={->,shorten <=1.5pt,shorten >=1.5pt}}
\scriptsize
\node[depot node](0) at (11.5mm,11.0mm){0};
\node[main node](1) at (18.0mm,10.0mm){1};
\node[main node](2) at (18.5mm,3.0mm){2};
\node[main node](3) at (21.5mm,17.0mm){3};
\node[main node](4) at (0.5mm,8.0mm){4};
\node[main node](5) at (8.5mm,1.0mm){5};
\node[main node](6) at (2.0mm,15.0mm){6};
\node[main node](7) at (14.0mm,17.0mm){7};
\node[main node](8) at (19.2mm,21.8mm){8};
\node[main node](9) at (6.1mm,22.0mm){9};
\Edge[style={bend right=15}](0)(8)
\Edge[style={bend right=7}](8)(7)
\Edge[style={bend right=7}](7)(9)
\Edge[style={bend right=7}](9)(5)
\Edge[style={bend left=7}](5)(6)
\Edge[style={bend left=0}](6)(4)
\Edge[style={bend right=7}](4)(1)
\Edge[style={bend left=7}](1)(2)
\Edge[style={bend right=7}](2)(3)
\Edge[style={bend left=7}](3)(0)
\end{tikzpicture}%
\caption{Solution before perturbation}%
\label{figure:init-perturb}%
\end{subfigure}%

\begin{subfigure}{.315\textwidth}%
\begin{tikzpicture}[>=latex',framed,scale=2,style={
main node/.style={circle,draw,minimum size=17,inner sep=1pt,thick},
depot node/.style={rectangle,draw,minimum size=15,inner sep=1pt,thick},
feat1 node/.style={circle,fill=gray!80,draw,minimum size=17,inner sep=1pt,thick},
feat2 node/.style={circle,fill=gray!20,draw,minimum size=17,inner sep=1pt,thick}}]
\SetUpEdge[lw = 0.75pt,color = black,labelcolor = white]
\tikzset{EdgeStyle/.style={->,shorten <=1.5pt,shorten >=1.5pt}}
\scriptsize
\node[depot node](0) at (11.5mm,11.0mm){0};
\node[main node](1) at (18.0mm,10.0mm){1};
\node[main node](2) at (18.5mm,3.0mm){2};
\node[main node](3) at (21.5mm,17.0mm){3};
\node[main node](4) at (0.5mm,8.0mm){4};
\node[main node](5) at (8.5mm,1.0mm){5};
\node[main node](6) at (2.0mm,15.0mm){6};
\node[feat1 node](7) at (14.0mm,17.0mm){7};
\node[main node](8) at (19.2mm,21.8mm){8};
\node[main node](9) at (6.1mm,22.0mm){9};
\Edge[style={bend right=20}](0)(8)
\Edge[style={bend right=7}](8)(7)
\Edge[style={bend right=7}](7)(9)
\Edge[style={bend right=7}](9)(5)
\Edge[style={bend left=7}](5)(6)
\Edge[style={bend right=0}](6)(4)
\Edge[style={bend right=7}](4)(1)
\Edge[style={bend left=7}](1)(2)
\Edge[style={bend right=7}](2)(3)
\Edge[style={bend right=7}](3)(7)
\Edge[style={bend right=7}](7)(0)
\end{tikzpicture}%
\caption{AddBuffer}%
\label{figure:addbuffer}%
\end{subfigure}%
\begin{subfigure}{.315\textwidth}%
\begin{tikzpicture}[>=latex',framed,scale=2,style={
main node/.style={circle,draw,minimum size=17,inner sep=1pt,thick},
depot node/.style={rectangle,draw,minimum size=15,inner sep=1pt,thick},
feat1 node/.style={circle,fill=gray!80,draw,minimum size=17,inner sep=1pt,thick},
feat2 node/.style={circle,fill=gray!40,draw,minimum size=17,inner sep=1pt,thick}}]
\SetUpEdge[lw = 0.75pt,color = black,labelcolor = white]
\tikzset{EdgeStyle/.style={->,shorten <=1.5pt,shorten >=1.5pt}}
\scriptsize
\node[depot node](0) at (11.5mm,11.0mm){0};
\node[main node](1) at (18.0mm,10.0mm){1};
\node[feat1 node](2) at (18.5mm,3.0mm){2};
\node[main node](3) at (21.5mm,17.0mm){3};
\node[main node](4) at (0.5mm,8.0mm){4};
\node[main node](5) at (8.5mm,1.0mm){5};
\node[feat1 node](6) at (2.0mm,15.0mm){6};
\node[main node](7) at (14.0mm,17.0mm){7};
\node[feat1 node](8) at (19.2mm,21.8mm){8};
\node[main node](9) at (6.1mm,22.0mm){9};
\Edge[style={bend right=15}](0)(8)
\Edge[style={bend right=7}](8)(7)
\Edge[style={bend right=7}](7)(9)
\Edge[style={bend right=7}](9)(5)
\Edge[style={bend left=7}](5)(6)
\Edge[style={bend right=0}](6)(4)
\Edge[style={bend right=10}](4)(1)
\Edge[style={bend right=0}](1)(2)
\Edge[style={bend right=7}](2)(3)
\Edge[style={bend right=7}](3)(8)
\Edge[style={bend right=7}](8)(6)
\Edge[style={bend right=7}](6)(2)
\Edge[style={bend right=0}](2)(0)
\end{tikzpicture}%
\caption{AddStations}%
\label{figure:addrandom}%
\end{subfigure}%
\begin{subfigure}{.315\textwidth}%
\begin{tikzpicture}[>=latex',framed,scale=2,style={
main node/.style={circle,draw,minimum size=17,inner sep=1pt,thick},
depot node/.style={rectangle,draw,minimum size=15,inner sep=1pt,thick},
feat1 node/.style={circle,fill=gray!80,draw,minimum size=17,inner sep=1pt,thick},
feat2 node/.style={circle,fill=gray!20,draw,minimum size=17,inner sep=1pt,thick}}]
\SetUpEdge[lw = 0.75pt,color = black,labelcolor = white]
\tikzset{EdgeStyle/.style={->,shorten <=1.5pt,shorten >=1.5pt}}
\scriptsize
\node[depot node](0) at (11.5mm,11.0mm){0};
\node[feat1 node](1) at (18.0mm,10.0mm){1};
\node[main node](2) at (18.5mm,3.0mm){2};
\node[main node](3) at (21.5mm,17.0mm){3};
\node[feat1 node](4) at (0.5mm,8.0mm){4};
\node[main node](5) at (8.5mm,1.0mm){5};
\node[feat1 node](6) at (2.0mm,15.0mm){6};
\node[feat2 node](7) at (14.0mm,17.0mm){7};
\node[main node](8) at (19.2mm,21.8mm){8};
\node[feat2 node](9) at (6.1mm,22.0mm){9};
\Edge[style={bend right=15}](0)(8)
\Edge[style={bend right=7}](8)(6)
\Edge[style={bend right=0}](6)(4)
\Edge[style={bend right=7}](4)(1)
\Edge[style={bend left=7}](1)(5)
\Edge[style={bend left=27}](5)(7)
\Edge[style={bend right=7}](7)(9)
\Edge[style={bend right=27}](9)(2)
\Edge[style={bend right=7}](2)(3)
\Edge[style={bend left=7}](3)(0)
\end{tikzpicture}%
\caption{Double-Bridge}%
\label{figure:double-bridge}%
\end{subfigure}%
\caption{Example regarding the application of perturbation mechanisms}%
\label{figure:perturbs}%
\end{figure}

Figure \ref{figure:perturbs} shows an example of perturbations applied over a (supposedly) local optimal solution. In Figure \ref{figure:double-bridge}, a Double-Bridge move is applied by interchanging  subsequence $6,4,1$ with subsequence $7,9$. Figure \ref{figure:addbuffer} shows the AddBuffer perturbation, when an additional visit to station $7$ is performed expecting it to act as a buffer. In Figure \ref{figure:addrandom}, the perturbation AddStations is applied by adding two random visits: one to station $8$ and another one to station $6$.

\section{Computational experiments}\label{sec:CompExp}

The ILS$_{\text{SBRP}}$ algorithm was coded in C++ (g++ 4.6.4) and the computational tests were carried on an Intel\textregistered Core\textsuperscript{TM} i7-3770 with 3.40 GHz and 16 GB of RAM running Ubuntu 14.04. Only a a single thread was used during the experiments.

\subsection{Instances}\label{sec:instances}

The benchmark instances used to test the proposed algorithm are those suggested by \citet{HerS04a}, which were originally created for the one-commodity pickup and delivery traveling salesman problem. The benchmark contains instances ranging from 20 to 500 customers (stations), and vehicle capacities ranging from 10 to 1000. For each pair of problem size and vehicle capacity, there are $10$ instances named from A to J and, for each vertex $i$, there is a demand $d_{i} \in [-10,10]$. \citet{chemla} and \citet{erdogan} only reported results for a subset of instances of the referred benchmark. Therefore, in order to compare our results with theirs, we  tested ILS$_{\text{SBRP}}$ for all instances considered in at least one of the two works (see Section \ref{sec:Results}).
Furthermore, to compute the initial and final targets as well as the load capacity for each station, the same procedures adopted by such authors were employed: for each vertex $i$, $p_{i} = \alpha \times 10$, $p^{\prime}_{i} = \alpha \times (10 + d_{i})$, $q_{i} = \alpha \times 20$, where $\alpha$ is an input parameter, and experiments were conducted with $\alpha = 1$ and $\alpha = 3$. In order to properly compare our results with those in \citet{chemla} and \citet{erdogan}, we adopted their same convention  
of rounding down the values of the cost matrix to the nearest integer (floor), although we noticed that this can cause slight violations of the triangle inequality. 



\subsection{Impact of the perturbation mechanisms}\label{sec:perturbanalisys}

In this section we are interested in evaluating the impact of the perturbation mechanisms described in Section \ref{sec:Perturb}, that is, AddBuffer ($P^{(1)}$), AddStations ($P^{(2)}$), Double-Bridge ($P^{(3)}$), and Suppression ($P^{(4)}$). In view of this, we selected a subset of 30 challenging instances for performing the experiments. 
These instances were chosen according to the largest gap values with respect to the lower bounds reported in \citet{erdogan}.
We ran ILS$_{\text{SBRP}}$ 10 times for each of the 30 instances considering all possible combinations of perturbations. For this testing we arbitrarily adopted $I_R = 10$ and $I_{ILS} = n$.


Table \ref{table:perturb-analysis} shows the impact of each combination over the average gaps (computed as $(UB-LB)/LB$) and CPU times required by ILS$_{\text{SBRP}}$ to run to completion. From the results obtained we believe that $P^{(2)}$  + $P^{(3)}$ + $P^{(4)}$ seems to offer a good compromise between solution quality (average gap $<$ 1\%) and CPU time. Hence, we herein decided to adopt this configuration for the perturbation mechanisms.  

{\scriptsize
    \begin{table}[ht!]
	\caption{Impact of different combinations of the perturbation mechanisms}\label{table:perturb-analysis}
	\centering
	\begin{tabular}{ccc}
	    \hline 
	    Perturbations & Avg. & Time \\ 
	    used & gap (\%) & (s) \\
	    \hline
	    $P^{(1)}$       & 2.43 & 244.40\\
	    $P^{(2)}$      & 1.08 & 773.49\\
	    $P^{(3)}$       & 1.04 & 498.24\\
	    $P^{(4)}$     & 1.56 & 249.88\\
	    $P^{(1)}$  + $P^{(2)}$     & 1.27 & 506.45\\
	    $P^{(1)}$  + $P^{(2)}$ + $P^{(3)}$  & 1.05 & 514.41\\
	    $P^{(1)}$  + $P^{(2)}$ + $P^{(3)}$  + $P^{(4)}$ & 1.11 & 431.43\\
	    $P^{(1)}$  + $P^{(2)}$ + $P^{(4)}$  & 1.25 & 417.30\\
	    $P^{(1)}$  + $P^{(3)}$     & 1.21 & 371.44\\
	    $P^{(1)}$  + $P^{(3)}$ + $P^{(4)}$  & 1.30 & 331.70\\
	    $P^{(1)}$  + $P^{(4)}$    & 1.78 & 248.69\\
	    $P^{(2)}$ + $P^{(3)}$    & 0.91 & 630.29\\
	    $P^{(2)}$  + $P^{(3)}$ + $P^{(4)}$  & 0.99 & 505.70\\
	    $P^{(2)}$  + $P^{(4)}$    & 1.08 & 503.75\\
	    $P^{(3)}$ + $P^{(4)}$    & 1.29 & 341.32\\
	    \hline
	\end{tabular}
    \end{table}
}


\subsection{Parameter tuning}\label{sec:parametertuning}


The main ILS$_{\text{SBRP}}$ parameters to be calibrated are the number of restarts ($I_R$) and the maximum number of consecutive ILS iterations without improvement over the current local optimal solution ($I_{ILS}$). Here we set $I_R = 10$, as in \citet{silva:2015}, where the authors put forward a multi-start ILS that was capable of obtaining state-of-the-art results for the split-delivery VRP.

In previous works, such as those mentioned in Section \ref{sec:Introduction}, the value of $I_{ILS}$ was tuned based on the instance size. In VRPs (e.g., \cite{Pennaetal2013, silva:2015,subramanian2012heuristic}, and \cite{Vidaletal2015}), this parameter is usually set as a function of the number of customers and vehicles. In TSP-like (or single machine scheduling) problems (e.g., \citealp{Silvaetal2012, SubramanianBattarra2013}; and \citealp{Subramanianetal2014}), $I_{ILS}$ was set only as a function of the number of customers (or jobs). We decided to use the same rationale as in \citet{Silvaetal2012}, by setting $I_{ILS} = \max\{I_{min}, \beta \times n\}$, where $I_{min}$ and  $\beta$ are input parameters. For the latter we set  $\beta = 4$, as in \citet{Subramanianetal2014}. Note that $I_{min}$ is more important for small size instances and its role is to prevent low values for $I_{ILS}$, which in this case may lead to an insufficient number of ILS iterations required for obtaining high quality (or even optimal) solutions. We then tested several values for $I_{min}$, more specifically, $100$, $120$, $140$, $160$, and $180$. For each of them, we ran ILS$_{\text{SBRP}}$ 10 times for all instances containing $20$ and $30$ stations. The average results obtained suggest that $I_{min} = 160$ seems to provide a good compromise between solution quality and CPU time, since the algorithm managed to find almost all best known solutions in a relatively short amount of time when using this value.  Therefore, we set $I_{ILS} = \max\{160, 4 \times n\}$.

\subsection{Comparison with the literature}\label{sec:Results}

ILS$_{\text{SBRP}}$ was executed 10 times for each instance with a time limit of 1 hour. The results found by our algorithm are compared with the upper bounds determined by two versions of the tabu search heuristic of \citet{chemla}. The first one  (TS1) starts from an initial solution generated by a greedy procedure, while the second version (TS2) receives the solution produced by their branch-and-cut algorithm (over a relaxation of the problem) as initial solution. Detailed results of TS1 and TS2 are available at \url{http://cermics.enpc.fr/~meuniefr/SVOCPDP.html}. According to the authors, it should be noted that UB1 is the best solution found by TS1 and UB2 is the best solution found considering both TS1 and TS2. A comparison is also performed with the lower and upper bounds obtained by the exact Branch-and-cut algorithm of \citet{erdogan}. Regarding the benchmark instances, \citet{chemla} considered $n \in \{20, 40, 60, 100\}$ and $Q \in \{10, 30, 45, 1000\}$, whereas \citet{erdogan} considered $n \in \{20, 30, 40, 50, 60\}$ and $Q \in \{10,15,20,25,30,35,40,45,1000\}$. 

\citet{chemla} ran their experiments on an AMD Athlon 5600+ 2.8 GHz with 16 GB of RAM, while \citet{erdogan} performed their testing on an Intel i7 3.60 GHz and 8 GB of RAM. On the one hand, because the hardware performance of the first appears to be quite inferior to the second, as well as to our intel i7 3.40 GHz, we decided to estimate an approximation factor based on the single thread rating values reported in \url{https://www.cpubenchmark.net/compare.php?cmp[]=86&cmp[]=896}, so as to better compare the runtime performance of the methods. According to the referred website,  the AMD Athlon 5600+ 2.8 GHz  is roughly 2.43 times slower than our processor. We thus report the original CPU time values of \citet{chemla} divided by a factor of 2.43. On the other hand, since the machine used in \cite{erdogan} is rather equivalent to ours, perhaps even slightly faster, we decided to consider the original runtime values reported by the authors.

\subsubsection{Results for instances with up to 60 stations}

The aggregate average results for instances containing 20, 30, 40, 50, and 60 stations are reported in Tables \ref{tab:alpha1} and \ref{tab:alpha3}, where \textbf{Instance group} denotes the set of 10 instances of a particular group (for example, group n20q10 contains 10 instances with $n = 20$ and $Q = 10$); \textbf{UB1 Gap (\%)}, \textbf{UB2 Gap (\%)}, and \textbf{Gap (\%)} correspond to the gap between UB1, UB2, and the upper bound found by \citet{erdogan}, respectively, and the lower bound reported in \cite{erdogan}; \textbf{Time (s)}, \textbf{UB1 Time (s)}, and \textbf{UB2 Time (s)} indicate, respectively, the CPU time in seconds spent by \citet{erdogan}, TS1, and TS2, where the last two are scaled to our processor as mentioned above; \textbf{Avg. Gap (\%)} and \textbf{Best Gap (\%)} are the gaps of the average solution and the best solution, respectively, found by ILS$_{\text{SBRP}}$ over the 10 runs with respect to the lower bounds in \cite{erdogan}; \textbf{Avg. Time (s)} is the average CPU time in seconds spent by ILS$_{\text{SBRP}}$ to completion over the 10 runs; \textbf{Avg. TT$_{\text{UB2}}$ (s)} denotes the average time over the 10 runs to find or improve the best heuristic solution found in \citet{chemla} (UB2); and \textbf{Avg. NV} is the average number of visits of the final solutions found by ILS$_{\text{SBRP}}$. Detailed results are reported in Appendix \ref{app:results}, including the best solution found when considering all experiments (\textbf{Best of all exp.}).

\begin{table}[!ht]
\caption{Aggregate average results per instance group for $n \in \{20, 30, 40, 50, 60\}$ and $\alpha = 1$}
\centering
{\footnotesize
\begin{tabular}{lccccccccccccc}
\hline\noalign{\smallskip}
& \multicolumn{2}{c}{\textbf{Erdo\v{g}an}} &&\multicolumn{4}{c}{\textbf{Chemla}} &&\\
& \multicolumn{2}{c}{\textbf{et al. 2015}} &&\multicolumn{4}{c}{\textbf{et al. 2013b}} &&\multicolumn{5}{c}{\up{\textbf{ILS$_{\text{SBRP}}$}}} \\
\cline{2-3}
\cline{5-8}
\cline{10-14}	\textbf{Instance}&&&&\textbf{UB1}&\textbf{UB1}&\textbf{UB2}&\textbf{UB2}&&\textbf{Avg.}&\textbf{Best}&\textbf{Avg.}&\textbf{Avg.}\\
\multicolumn{1}{c}   {\textbf{group}}&\up{\textbf{Gap}}&\up{\textbf{Time}}&&\textbf{Gap}&\textbf{Time}&\textbf{Gap}&\textbf{Time}&&\textbf{Gap}&\textbf{Gap}&\textbf{Time}&\textbf{TT$_{\text{UB2}}$}&\up{\textbf{Avg.}}\\
&\up{\textbf{(\%)}}&\up{\textbf{(s)}}&&\textbf{(\%)}&\textbf{(s)}&\textbf{(\%)}&\textbf{(s)}&&\textbf{(\%)}&\textbf{(\%)}&\textbf{(s)}&\textbf{(s)}&\textbf{\up{NV}}\\
\hline
n20q10&0.00&0.35&  &2.57&2.75&0.00&3.53& &0.06&0.00&5.84&0.38&20.10\\
n20q15&0.00&0.30& &-&-&-&-&  &0.00&0.00&2.13&-&19.19\\
n20q20&0.00&0.13& &-&-&-&-&  &0.00&0.00&0.94&-&18.65\\
n20q25&0.00&0.15& &-&-&-&-&  &0.00&0.00&0.63&-&18.62\\
n20q30&0.00&1.96&  &2.47&2.22&0.39&2.38& &0.00&0.00&0.62&0.01&18.66\\
n20q35&0.00&1.12& &-&-&-&-&  &0.00&0.00&0.46&-&18.70\\
n20q40&0.00&1.22& &-&-&-&-&  &0.00&0.00&0.46&-&18.65\\
n20q45&0.00&1.13&  &1.52&2.71&0.01&2.83& &0.00&0.00&0.45&0.01&18.62\\
n20q1000&0.00&0.83&  &2.43&2.92&0.00&3.04& &0.00&0.00&0.45&0.01&18.62\\
\hline
\multicolumn{1}{c}{Avg.}&0.00&0.80&&2.25&2.65&0.10&2.95&&0.01&0.00&1.33&0.10&18.87\\
\hline
n30q10&0.00&6.22&  &-&-&-&-&&0.02&0.00&28.08&-&30.14\\
n30q15&0.00&3.87&  &-&-&-&-&&0.12&0.00&10.29&-&28.49\\
n30q20&0.00&163.59&  &-&-&-&-&&0.02&0.02&5.06&-&27.84\\
n30q25&0.00&5.61&  &-&-&-&-&&0.00&0.00&2.36&-&27.53\\
n30q30&0.00&82.20&  &-&-&-&-&&0.00&0.00&1.85&-&27.53\\
n30q35&0.00&293.27&  &-&-&-&-&&0.02&0.00&1.89&-&27.70\\
n30q40&0.00&584.62&  &-&-&-&-&&0.00&0.00&1.61&-&27.62\\
n30q45&0.00&221.69&  &-&-&-&-&&0.01&0.00&1.47&-&27.62\\
n30q1000&0.00&190.20&  &-&-&-&-&&0.00&0.00&1.42&-&27.64\\
\hline
\multicolumn{1}{c}{Avg.}&0.00&172.36& &-&-&-&-&&0.02&0.00&6.00&-&28.01\\
\hline
n40q10&0.00&124.80&  &3.56&151.92&0.09&1752.75& &0.09&0.04&56.89&14.02&39.76\\
n40q15&0.00&25.55&  &-&-&-&-& &0.01&0.00&21.78&-&37.53\\
n40q20&0.00&14.72&  &-&-&-&-& &0.01&0.00&9.33&-&37.01\\
n40q25&0.03&723.88&  &-&-&-&-& &0.04&0.03&5.79&-&36.83\\
n40q30&0.00&36.56&  &4.05&92.72&0.00&93.87& &0.00&0.00&4.43&0.06&36.79\\
n40q35&0.00&38.66&  &-&-&-&-& &0.00&0.00&3.36&-&36.77\\
n40q40&0.00&70.65&  &-&-&-&-& &0.00&0.00&3.13&-&36.79\\
n40q45&0.00&74.28&  &4.69&93.63&0.61&94.82& &0.00&0.00&2.99&0.04&36.83\\
n40q1000&0.00&70.17&  &5.07&112.48&0.77&114.54& &0.00&0.00&2.95&0.03&36.82\\
\hline
\multicolumn{1}{c}{Avg.}&0.00&131.03&&4.34&112.69&0.37&514.00&&0.02&0.01&12.29&3.54&37.24\\
\hline
n50q10&0.79&1198.48&  &-&-&-&-&&0.30&0.23&210.98&-&49.64\\
n50q15&0.43&1970.12&  &-&-&-&-&&0.29&0.23&75.07&-&46.71\\
n50q20&0.00&295.45&  &-&-&-&-&&0.05&0.00&35.85&-&46.11\\
n50q25&0.00&272.82&  &-&-&-&-&&0.00&0.00&25.52&-&45.71\\
n50q30&0.00&177.40&  &-&-&-&-&&0.00&0.00&13.18&-&45.47\\
n50q35&0.24&1461.09&  &-&-&-&-&&0.16&0.16&10.61&-&45.63\\
n50q40&0.01&1408.94&  &-&-&-&-&&0.01&0.01&8.83&-&45.48\\
n50q45&0.00&1221.33&  &-&-&-&-&&0.00&0.00&8.35&-&45.41\\
n50q1000&0.11&1909.76&  &-&-&-&-&&0.09&0.09&6.70&-&45.44\\
\hline
\multicolumn{1}{c}{Avg.}&0.18&1101.71& &-&-&-&-&&0.10&0.08&43.90&-&46.18\\
\hline
n60q10&1.24&3924.62&  &13.62&412.18&2.57&4533.96& &0.67&0.57&419.95&35.15&60.21\\
n60q15&0.51&1957.50& &-&-&-&-& &0.30&0.27&140.29&-&56.48\\
n60q20&0.00&1285.03& &-&-&-&-& &0.00&0.00&72.99&-&55.65\\
n60q25&0.07&943.42& &-&-&-&-& &0.07&0.07&44.32&-&55.13\\
n60q30&0.13&1252.65&  &7.17&416.50&0.57&911.58& &0.08&0.07&29.63&2.46&55.22\\
n60q35&0.13&1096.98& &-&-&-&-& &0.06&0.05&22.52&-&55.01\\
n60q40&0.22&2607.91& &-&-&-&-& &0.19&0.19&21.10&-&55.33\\
n60q45&0.15&2795.20&  &7.11&410.29&1.78&427.01& &0.16&0.15&19.85&0.43&55.42\\
n60q1000&0.18&2816.42&  &7.14&413.87&1.76&515.42& &0.18&0.18&16.28&0.49&55.31\\
\hline
\multicolumn{1}{c}{Avg.}&0.29&2075.52&&8.76&413.21&1.67&1596.99&&0.19&0.17&87.44&9.63&55.97\\
\hline
\end{tabular}}
\label{tab:alpha1}
\end{table}

From Table \ref{tab:alpha1}, it can be observed that the quality of  the solutions found by ILS$_{\text{SBRP}}$, as well as those obtained by the algorithm of \citet{erdogan}, are visibly superior than the ones determined by the tabu searches of \citet{chemla}, especially TS1. Such superiority becomes even more prominent for $\alpha = 3$, as presented in Table \ref{tab:alpha3}. Also, the average CPU times spent by ILS$_{\text{SBRP}}$ to find or improve the best solutions reported by \citet{chemla} (UB2) are rather small in most cases, sometimes only a matter of relatively few seconds, except for the instance group n40q10 when $\alpha = 3$, where the proposed algorithm required more CPU time.     

In addition, assuming the same values for the demands, the smaller the vehicle capacity, the larger the relative number of visits. This increases the size of the tour, thus affecting the number of operations performed during the local search, and possibly the number of ILS iterations, as more moves are required to be evaluated.

\begin{table}[!ht]
\caption{Aggregate average results per instance group for $n \in \{20, 30, 40, 50, 60\}$ and $\alpha = 3$}
\centering
{\small
\renewcommand{\tabcolsep}{0.173cm}
\begin{tabular}{lccccccccccccc}
\hline\noalign{\smallskip}
& \multicolumn{2}{c}{\textbf{Erdo\v{g}an}} 
&&\multicolumn{4}{c}{\textbf{Chemla}} &&\\
& \multicolumn{2}{c}{\textbf{et al. 2015}} 
&&\multicolumn{4}{c}{\textbf{et al. 2013b}} 
&&\multicolumn{5}{c}{\up{\textbf{ILS$_{\text{SBRP}}$}}} \\
\cline{2-3}
\cline{5-8}
\cline{10-14}
\textbf{Instance}&&&&\textbf{UB1}&\textbf{UB1}&\textbf{UB2}&\textbf{UB2}&&\textbf{Avg.}&\textbf{Best}&\textbf{Avg.}&\textbf{Avg.}\\
\multicolumn{1}{c}{\textbf{group}}&\up{\textbf{Gap}}&\up{\textbf{Time}}&&\textbf{Gap}&\textbf{Time}&\textbf{Gap}&\textbf{Time}&&\textbf{Gap}&\textbf{Gap}&\textbf{Time}&
\textbf{TT$_{\text{UB2}}$}&\up{\textbf{Avg.}}\\
&\up{\textbf{(\%)}}&\up{\textbf{(s)}}&&\textbf{(\%)}&\textbf{(s)}&\textbf{(\%)}&\textbf{(s)}&&\textbf{(\%)}&\textbf{(\%)}&\textbf{(s)}&\textbf{(s)}&\up{\textbf{NV}}\\
\hline
n20q10&0.00&0.48&  &0.73&40.71&0.00&51.72& &0.02&0.00&219.94&18.77&39.38\\
n20q15&0.00&0.32& &-&-&-&-&  &0.01&0.00&46.86&-&29.27\\
n20q20&0.00&0.37& &-&-&-&-&  &0.01&0.00&32.96&-&27.16\\
n20q25&0.00&0.39& &-&-&-&-&  &\textbf{0.00}&0.00&12.10&-&22.29\\
n20q30&0.00&0.35&  &3.93&3.29&0.00&4.44& &0.07&0.00&5.69&0.28&20.03\\
n20q35&0.00&0.33& &-&-&-&-&  &0.16&0.05&4.46&-&19.71\\
n20q40&0.00&0.28& &-&-&-&-&  &\textbf{0.00}&0.00&2.72&-&19.34\\
n20q45&0.00&0.31&  &3.39&2.67&0.00&3.62& &\textbf{0.00}&0.00&2.12&0.05&19.17\\
n20q1000&0.00&0.93&  &3.47&2.63&0.00&2.63& &\textbf{0.00}&0.00&0.45&0.01&18.60\\
\hline
\multicolumn{1}{c}{Avg.}&0.00&0.42&&2.88&12.33&0.00&15.60&&0.03&0.01&36.37&4.78&23.88\\
\hline
n30q10&0.00&153.85&  &-&-&-&-&&0.04&0.00&1115.36&-&57.39\\
n30q15&0.00&65.36&  &-&-&-&-&&0.03&0.00&206.32&-&42.47\\
n30q20&0.00&8.16&  &-&-&-&-&&0.03&0.00&133.46&-&39.38\\
n30q25&0.00&10.27&  &-&-&-&-&&0.10&0.03&65.47&-&33.97\\
n30q30&0.00&9.91&  &-&-&-&-&&0.02&0.00&28.78&-&30.12\\
n30q35&0.00&9.37&  &-&-&-&-&&0.22&0.00&24.29&-&29.83\\
n30q40&0.00&5.92&  &-&-&-&-&&0.08&0.05&11.85&-&28.32\\
n30q45&0.00&2.96&  &-&-&-&-&&0.14&0.00&9.96&-&28.49\\
n30q1000&0.00&221.30&  &-&-&-&-&&0.02&0.00&1.39&-&27.59\\
\hline
\multicolumn{1}{c}{Avg.}&0.00&54.12& &-&-&-&-&&0.08&0.01&177.43&-&35.28\\
\hline
n40q10&0.00&235.42&  &4.76&408.07&0.39&3983.91& &0.07&0.00&2619.68&457.03&73.67\\
n40q15&0.00&28.83&  &-&-&-&-& &0.05&0.01&390.64&-&53.45\\
n40q20&0.00&62.22&  &-&-&-&-& &0.01&0.00&331.38&-&50.29\\
n40q25&0.00&177.39&  &-&-&-&-& &0.06&0.01&136.27&-&43.57\\
n40q30&0.00&108.31&  &4.64&101.43&0.00&1524.21& &0.10&0.04&57.86&19.28&39.63\\
n40q35&0.00&304.70&  &-&-&-&-& &0.13&0.03&55.48&-&39.35\\
n40q40&0.00&21.68&  &-&-&-&-& &0.01&0.00&34.54&-&38.30\\
n40q45&0.00&25.74&  &5.14&113.43&0.00&219.71& &0.01&0.00&21.41&3.29&37.49\\
n40q1000&0.00&80.91&  &4.58&91.08&0.30&95.60& &\textbf{0.00}&0.00&2.97&0.03&36.81\\
\hline
\multicolumn{1}{c}{Avg.}&0.00&116.13&&4.78&178.50&0.17&1455.86&&0.05&0.01&405.58&119.91&45.84\\
\hline
n50q10&0.99&2693.41&  &-&-&-&-&&\textbf{0.33}&0.19&3579.49&-&95.30\\
n50q15&0.00&1702.17&  &-&-&-&-&&0.06&0.03&1503.56&-&69.44\\
n50q20&0.89&3085.39&  &-&-&-&-&&\textbf{0.53}&0.42&1374.12&-&64.04\\
n50q25&0.59&2024.55&  &-&-&-&-&&\textbf{0.34}&0.24&544.63&-&54.96\\
n50q30&0.46&1345.21&  &-&-&-&-&&\textbf{0.38}&0.29&215.59&-&49.75\\
n50q35&1.27&4212.45&  &-&-&-&-&&\textbf{0.79}&0.59&201.46&-&49.60\\
n50q40&0.45&3545.41&  &-&-&-&-&&\textbf{0.45}&0.39&143.50&-&48.49\\
n50q45&0.23&2057.76&  &-&-&-&-&&\textbf{0.23}&0.18&74.94&-&46.75\\
n50q1000&0.10&1938.98&  &-&-&-&-&&\textbf{0.09}&0.09&6.66&-&45.41\\
\hline
\multicolumn{1}{c}{Avg.}&0.55&2511.70& &-&-&-&-&&\textbf{0.36}&0.27&849.33&-&58.19\\
\hline
n60q10&2.83&3718.02&  &46.25&401.91&5.97&4524.10& &\textbf{0.63}&0.30&3602.77&39.70&115.50\\
n60q15&2.73&2932.58& &-&-&-&-& &\textbf{0.41}&0.36&2613.02&-&85.09\\
n60q20&0.39&3772.60& &-&-&-&-& &\textbf{0.26}&0.19&2619.83&-&78.76\\
n60q25&2.63&3636.12& &-&-&-&-& &\textbf{0.59}&0.46&1153.57&-&67.26\\
n60q30&2.88&4702.35&  &13.48&412.35&5.06&4533.51& &\textbf{0.74}&0.57&430.87&5.75&60.34\\
n60q35&2.42&4795.69& &-&-&-&-& &\textbf{0.87}&0.66&418.48&-&60.05\\
n60q40&0.46&3829.99& &-&-&-&-& &\textbf{0.37}&0.31&244.74&-&57.60\\
n60q45&0.26&2223.52&  &15.14&413.41&2.04&3840.61& &\textbf{0.26}&0.23&136.54&7.77&56.44\\
n60q1000&0.18&2940.39&  &7.47&414.81&1.94&483.46& &\textbf{0.18}&0.17&15.97&0.11&55.26\\
\hline
\multicolumn{1}{c}{Avg.}&1.64&3616.81&&20.59&410.62&3.75&3345.42& &\textbf{0.48}&0.36&1248.42&13.33&70.70\\
\hline
\end{tabular}}
\label{tab:alpha3}
\end{table}

\begin{figure}[!ht]
	\centering
	\begin{subfigure}{.500\textwidth}%
		\includegraphics[scale=0.6]{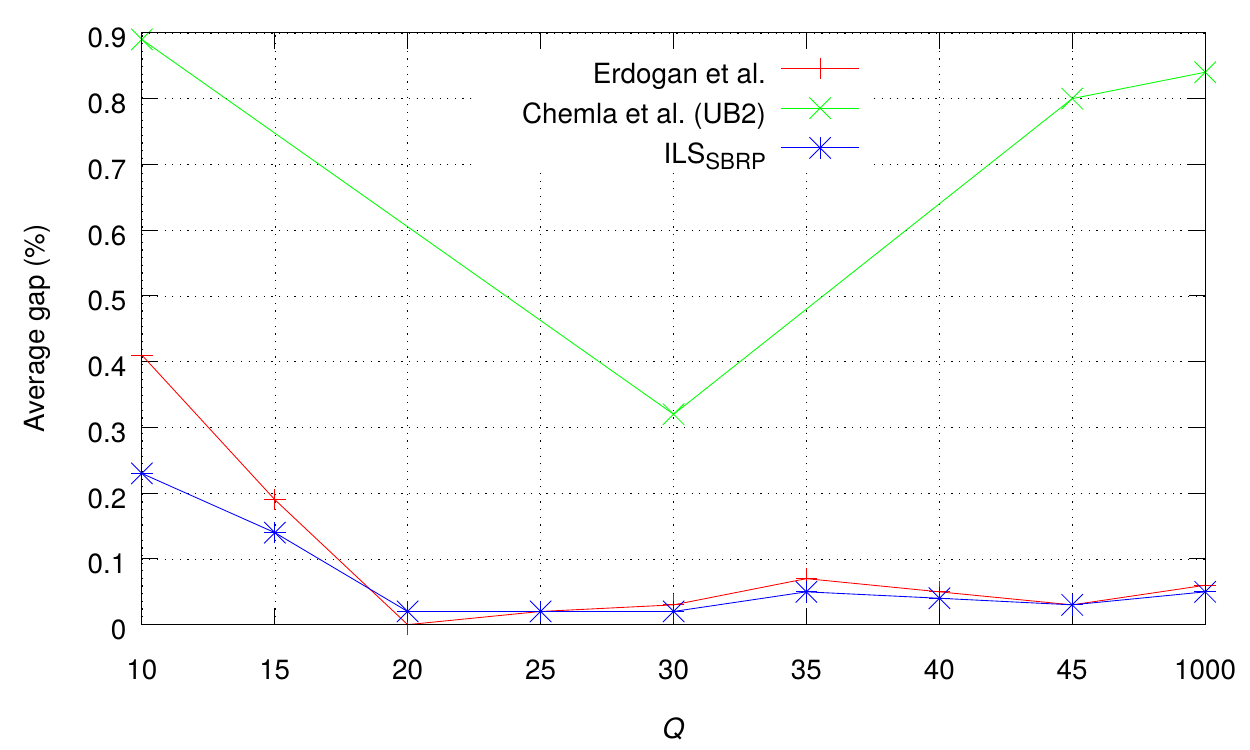}%
		\caption{Average gap (\%)}%
		\label{figure:avg-gap-a1-by-q}%
	\end{subfigure}%
	\begin{subfigure}{.500\textwidth}%
		\includegraphics[scale=0.6]{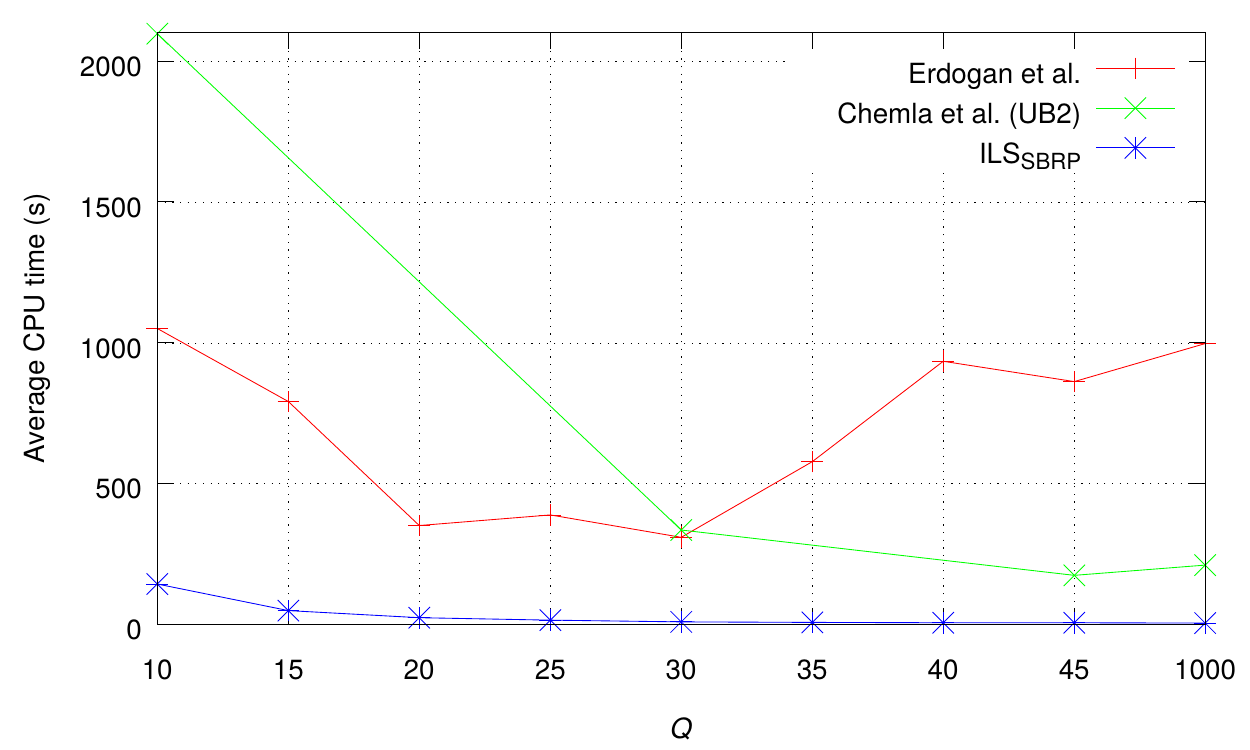}%
		\caption{Average CPU time (s)}%
		\label{figure:avg-times-a1-by-q}%
	\end{subfigure}%
	\caption{Average gap (\%) and average CPU time (s) per $Q$ and $\alpha = 1$}
	\label{figure:avg-gaps-times-a1-by-q}
\end{figure}

\begin{figure}[!ht]
	\centering
	\begin{subfigure}{.500\textwidth}%
		\includegraphics[scale=0.6]{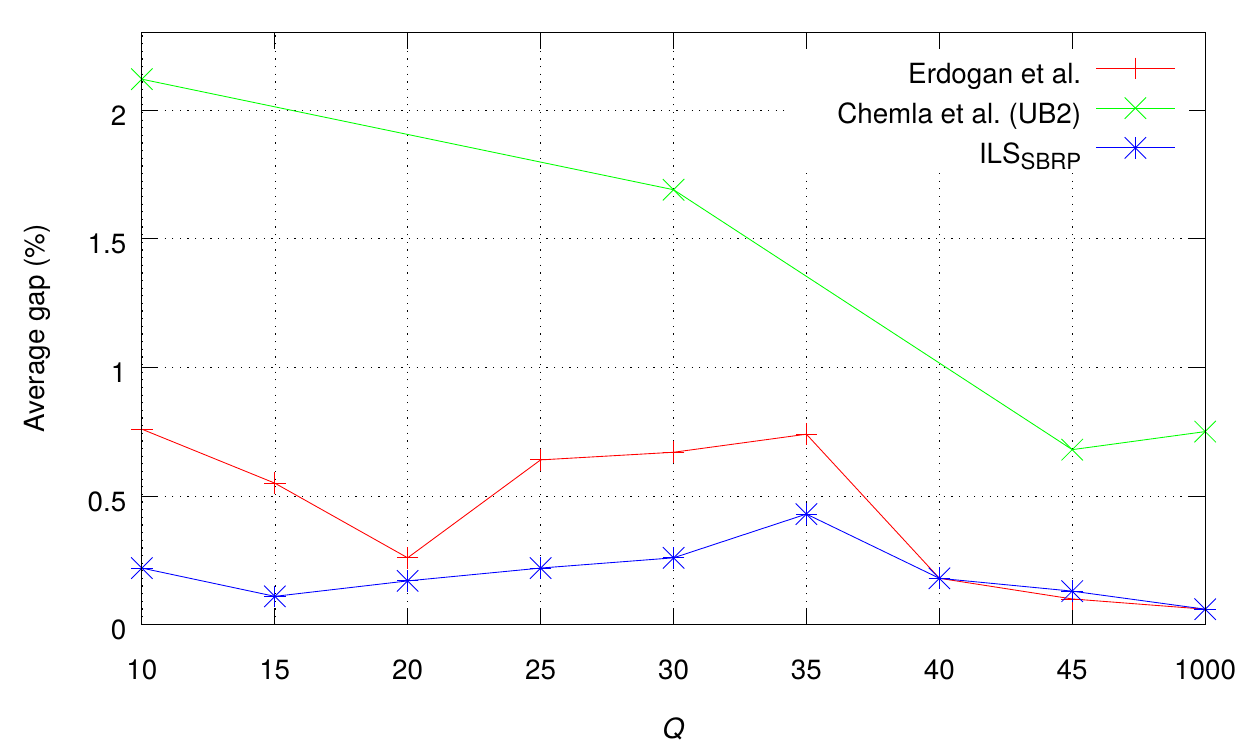}%
		\caption{Average gap (\%)}%
		\label{figure:avg-gap-a3-by-q}%
	\end{subfigure}%
	\begin{subfigure}{.500\textwidth}%
		\includegraphics[scale=0.6]{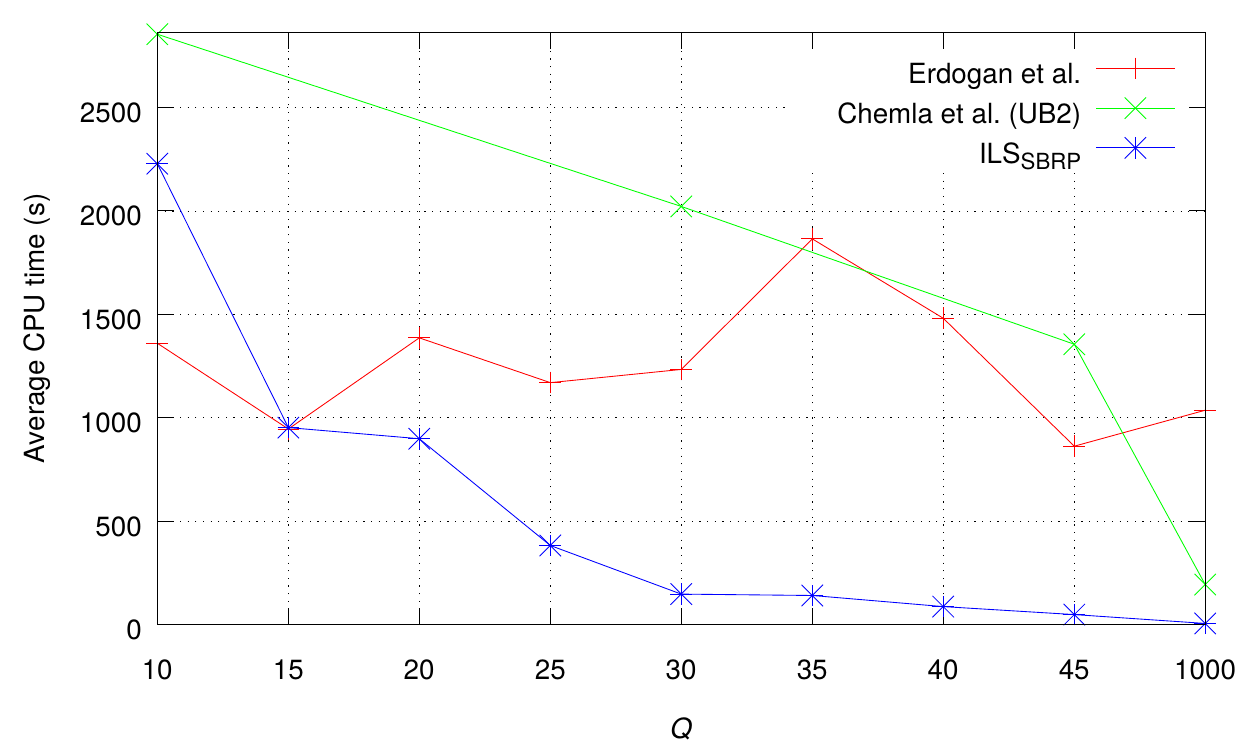}%
		\caption{Average CPU time (s)}%
		\label{figure:avg-times-a3-by-q}%
	\end{subfigure}%
	\caption{Average gap (\%) and average CPU time (s) per $Q$ and $\alpha = 3$}
	\label{figure:avg-gaps-times-a3-by-q}
\end{figure}

Figures \ref{figure:avg-gaps-times-a1-by-q} and \ref{figure:avg-gaps-times-a3-by-q} show how the average gaps and CPU times of each method vary according to the value of $Q$.  We omit the results of TS1 because the associated gaps are quite inferior when compared to those obtained by the other algorithms. While the average gaps of ILS$_{\text{SBRP}}$ tend to be larger for very small values of $Q$, the average CPU time decreases as the value of $Q$ increases, both for $\alpha = 1$ and $\alpha = 3$. A similar behavior regarding the CPU time performance can be observed for the heuristic of \citet{chemla}, as opposed to the algorithm of \citet{erdogan}, which does not seem to have a consistent pattern when considering this aspect.

\begin{figure}[!ht]
	\centering
	\begin{subfigure}{.500\textwidth}%
		\includegraphics[scale=0.6]{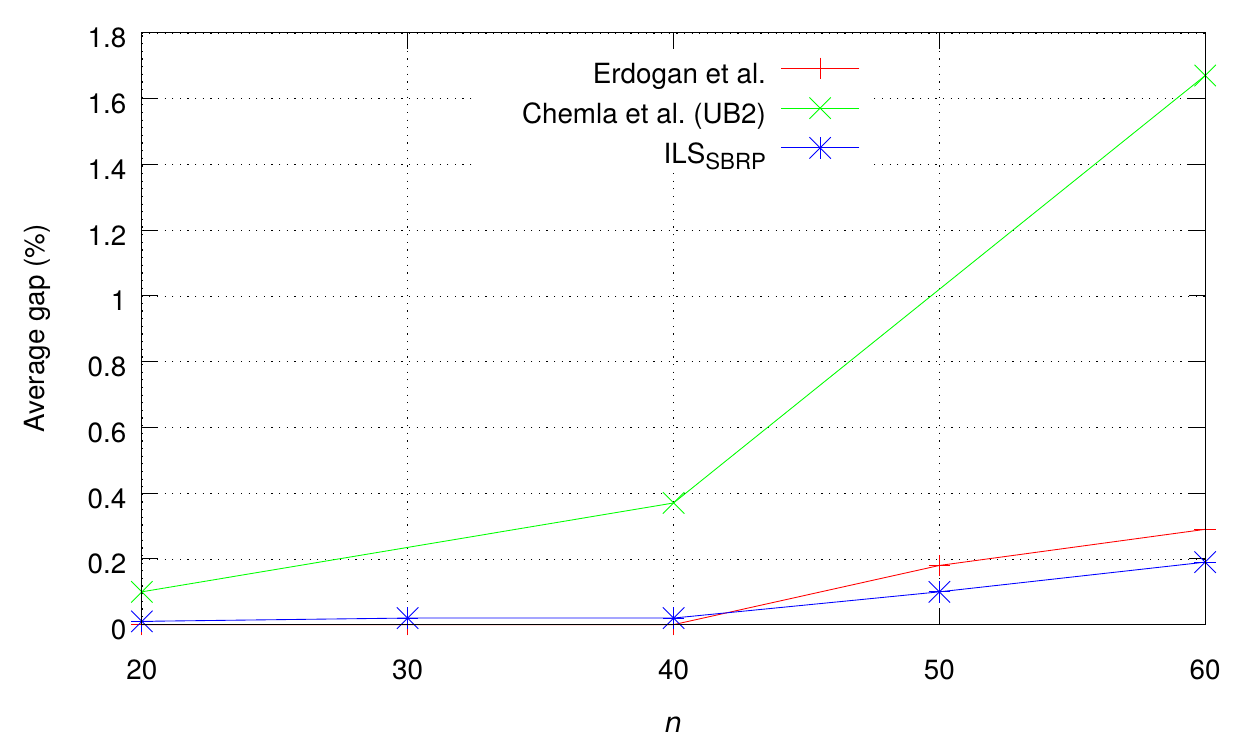}%
		\caption{Average gap (\%)}%
		\label{figure:avg-gap-a1-by-n}%
	\end{subfigure}%
	\begin{subfigure}{.500\textwidth}%
		\includegraphics[scale=0.6]{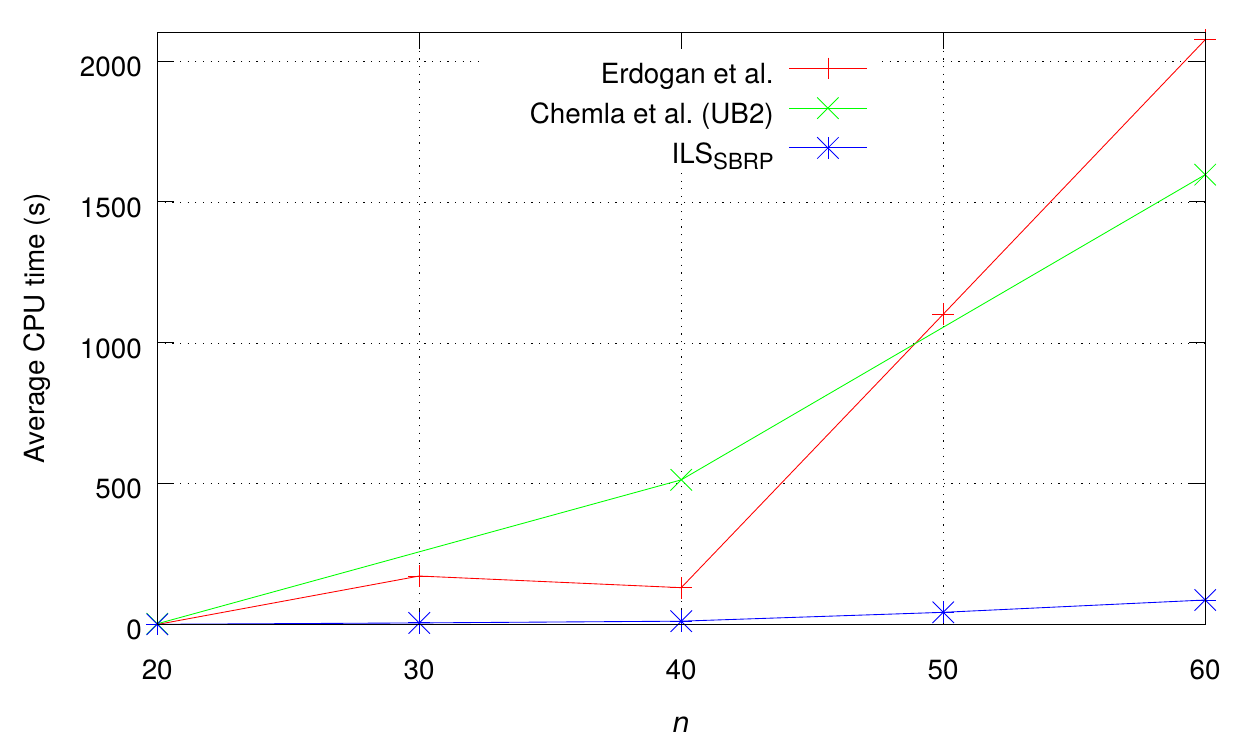}%
		\caption{Average CPU time (s)}%
		\label{figure:avg-times-a1-by-n}%
	\end{subfigure}%
	\caption{Average gap (\%) and average CPU time (s) per $n$ and $\alpha = 1$}
	\label{figure:avg-gaps-times-a1-by-n}
\end{figure}

\begin{figure}[!ht]
	\centering
	\begin{subfigure}{.500\textwidth}%
		\includegraphics[scale=0.6]{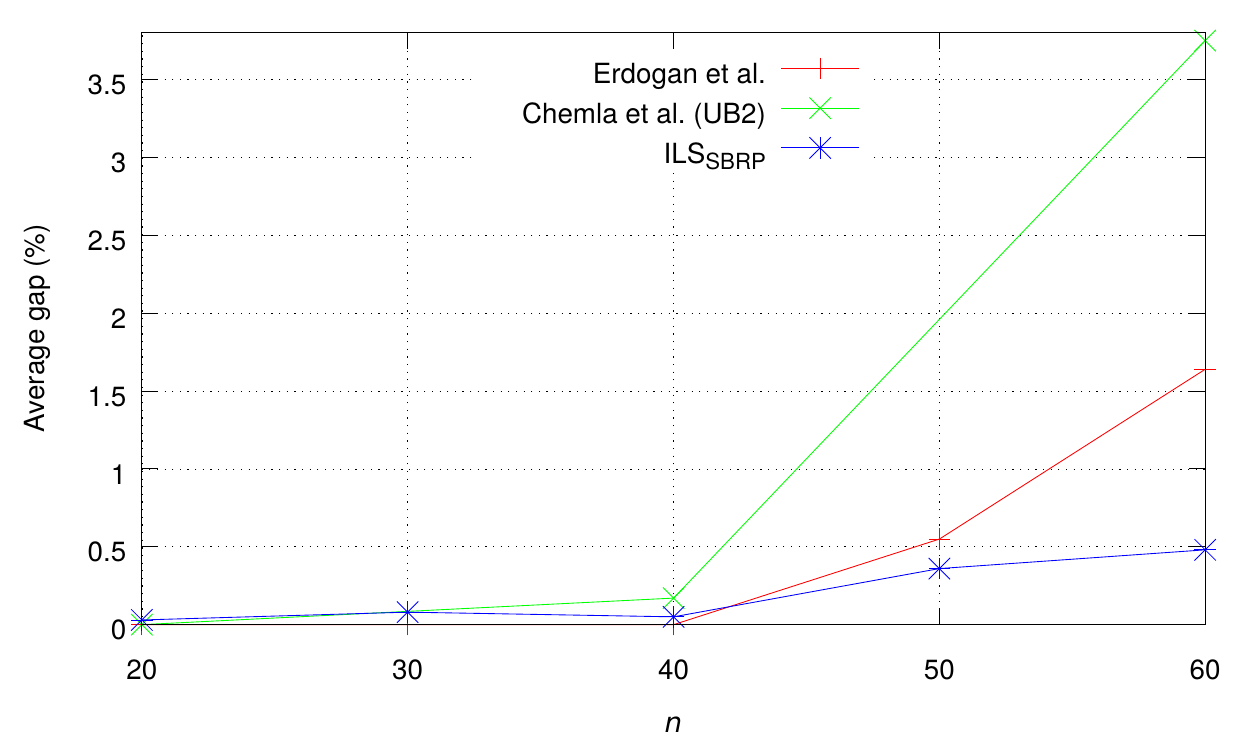}%
		\caption{Average gap (\%)}%
		\label{figure:avg-gap-a3-by-n}%
	\end{subfigure}%
	\begin{subfigure}{.500\textwidth}%
		\includegraphics[scale=0.6]{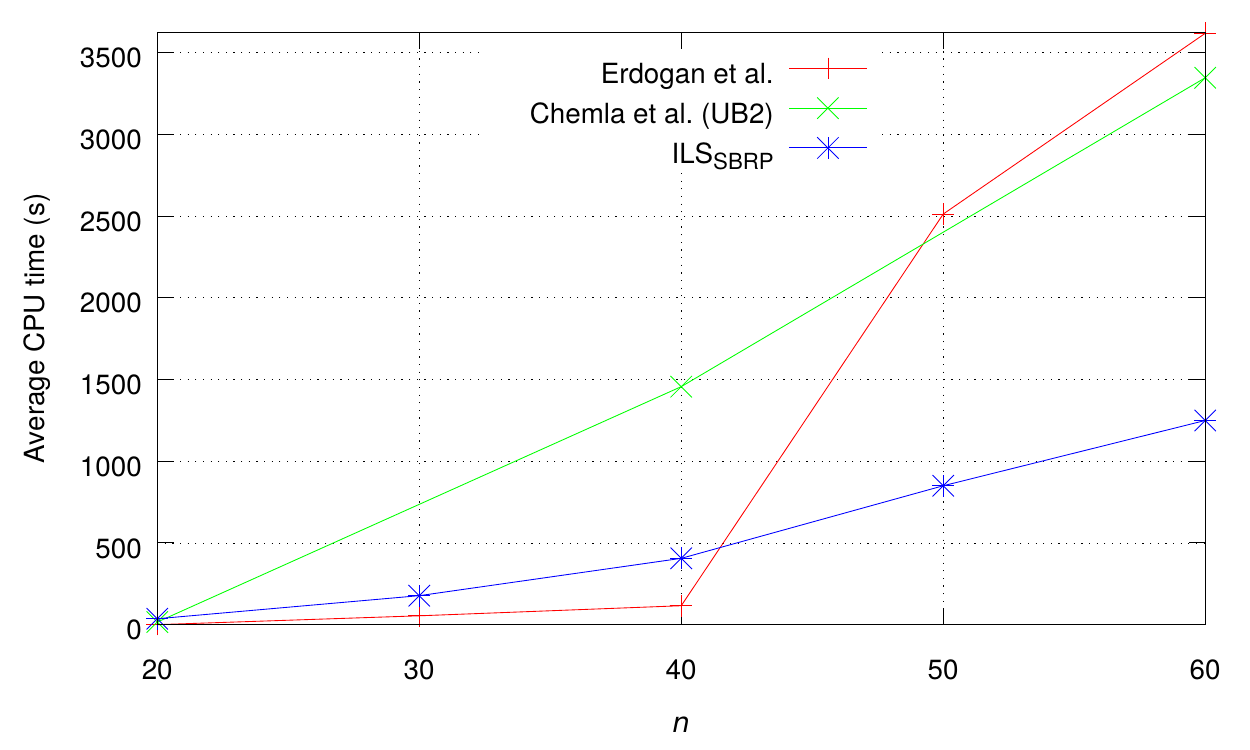}%
		\caption{Average CPU time (s)}%
		\label{figure:avg-times-a3-by-n}%
	\end{subfigure}%
	\caption{Average gap (\%) and average CPU time (s) per $n$ and $\alpha = 3$}
	\label{figure:avg-gaps-times-a3-by-n}
\end{figure}

Figures \ref{figure:avg-gaps-times-a1-by-n} and \ref{figure:avg-gaps-times-a3-by-n} illustrate the behavior of the average gaps and CPU times of the algorithms as the number of stations increases. Overall, the quality of the solutions found by ILS$_{\text{SBRP}}$ and by the algorithm of \citet{erdogan} are equivalent except for $n = 60$ and $\alpha = 3$, where the former clearly outperforms the latter. Moreover, there is a considerable increase on the CPU time for both methods from the literature for $n > 40$, in contrast to  ILS$_{\text{SBRP}}$, whose increase appears to be more moderate. However, this was somewhat expected since the  CPU effort of the algorithms of \citet{erdogan} and of \citet{chemla} tend to increase exponentially with increasing values of $n$.

Finally, Table \ref{tab:summary-a1-3} shows a summary of the best solutions found by the proposed algorithm, where \textbf{$\#$Opt} denotes the number of optimal solutions,  \textbf{$\#$Impr.} corresponds to the number of improved solutions, \textbf{$\#$Equal} indicates the number of equal solutions, and \textbf{$\#$Worse} is the number of worse solutions. The results of \citet{chemla} were not included in these tables because they are dominated by those obtained by our heuristic for all instances. For $\alpha = 1$, ILS$_{\text{SBRP}}$ found 419 of the 424 known optimal solutions and improved the result of 17 of the 26 instances that remain open. As for $\alpha = 3$, ILS$_{\text{SBRP}}$ achieved 377 out of 399 proven optimal solutions and improved the best known solution of 44 out of 51 open instances.

\begin{table}[!ht]
\caption{Summary of the performance of the best solutions aggregated by $n$}
\centering
{\small
\begin{tabular}{cccccccc}
\hline\noalign{\smallskip}
&& \multicolumn{1}{c}{\textbf{Erdo\v{g}an}} &&\\
&& \multicolumn{1}{c}{\textbf{et al. 2015}} &&\multicolumn{4}{c}{\up{\textbf{ILS$_{\text{SBRP}}$}}} \\
\cline{3-3}
\cline{5-8}
&\up{\textbf{\emph{n}}}&\textbf{$\#$Opt} &&\textbf{$\#$Opt} &  \textbf{$\#$Impr.}&\textbf{$\#$Equal}&\textbf{$\#$Worse} \\
\hline
\multirow{6}{*}{$\alpha=1$}&20&90&&90&0&90&0\\
&30&90&&89&0&89&1\\
&40&89&&87&0&88&2\\
&50&83&&83&5&85&0\\
&60&72&&70&12&76&2\\
\cline{2-8}
&\multicolumn{1}{c}{Total}&424&&419&17&428&5\\
\hline
\multirow{6}{*}{$\alpha=3$}&20&90&&89&0&89&1\\
&30&90&&87&0&87&3\\
&40&90&&83&0&83&7\\
&50&69&&63&17&67&6\\
&60&60&&55&27&58&5\\
\cline{2-8}
&\multicolumn{1}{c}{Total}&399&&377&44&384&22\\
\hline
\end{tabular}}
\label{tab:summary-a1-3}
\end{table}

\subsubsection{Results for instances with 100 stations}

Tables \ref{tab:n100-a1} and \ref{tab:n100-a3} illustrate the detailed results found by our algorithm and the tabu searches of \citet{chemla} for every instance containing 100 stations. In this case, the gaps are computed with respect to the lower bound reported in \cite{chemla}. The results obtained show that ILS$_{\text{SBRP}}$ clearly outperforms the methods from the literature, both in terms of solution quality and CPU time. In general, the proposed heuristic was capable of significantly improving the best known solution of all instances. 

In addition, when considering $\alpha = 1$, ILS$_{\text{SBRP}}$ required, on average, at most 8 seconds to find or improve the best results of \citet{chemla}. In some cases, such as those involving $Q \geq 30$, our algorithm spent, on average, only a fraction of a second to achieve a superior solution than the best one from the literature. For $\alpha = 3$, more time was required, on average, to accomplish the same purpose, but mostly for $Q = 10$.

\begin{table}[!ht]
\caption{Detailed results for $n = 100$ and $\alpha = 1$}
\centering
{\small
\renewcommand{\tabcolsep}{0.085cm}
{\renewcommand{\arraystretch}{1.1}
\begin{tabular}{lcccccccccccccc}
\hline\noalign{\smallskip}
& \multicolumn{6}{c}{\textbf{Chemla}} & & & & & &&\\
& \multicolumn{6}{c}{\textbf{et al. 2013b}} &&\multicolumn{7}{c}{\up{\textbf{ILS$_{\text{SBRP}}$}}} \\
\cline{2-7}
\cline{9-15}
\textbf{Instance} &&\textbf{UB1}&\textbf{UB1} &&\textbf{UB2}&\textbf{UB2} & &&\textbf{Avg.} &&\textbf{Best}&\textbf{Avg.}&\textbf{Avg.} \\
&\textbf{UB1}&\textbf{Gap}&\textbf{Time}&\textbf{UB2}&\textbf{Gap}&\textbf{Time}&&\up{\textbf{Avg.}}&\textbf{Gap}&\up{\textbf{Best}}&\textbf{Gap}&\textbf{Time}& \textbf{TT$_{\text{UB2}}$}&\up{\textbf{Avg.}} \\
& &\textbf{(\%)}&\textbf{(s)}&&\textbf{(\%)}&\textbf{(s)}&&\up{\textbf{Sol.}}& \textbf{(\%)}&\up{\textbf{Sol.}}&\textbf{(\%)}&\textbf{(s)}&\textbf{(s)}& \up{\textbf{NV}} \\
\hline
n100q10A&14921&45.58&681.15&13273&29.50&4805.44&&11283.20&10.09&11258&9.84&3558.39&3.23&97.30\\
n100q10B&17658&59.29&470.81&14981&35.14&4588.93&&12669.20&14.29&12609&13.75&3600.31&5.28&102.20\\
n100q10C&17138&44.75&617.06&15636&32.07&4758.60&&13251.00&11.92&13224&11.69&3600.47&4.72&100.70\\
n100q10D&19278&57.05&489.29&16586&35.12&4620.98&&13832.80&12.69&13783&12.29&3600.54&3.23&98.40\\
n100q10E&16867&69.73&368.51&12513&25.91&4485.81&&10974.90&10.44&10954&10.23&3385.98&6.10&105.70\\
n100q10F&14759&46.88&486.42&12621&25.60&4603.72&&11226.20&11.72&11191&11.37&3600.31&5.26&101.30\\
n100q10G&16772&63.48&380.01&13820&34.71&4498.96&&11186.60&9.04&11160&8.78&3347.66&2.40&100.10\\
n100q10H&15941&45.21&393.16&14863&35.39&4511.70&&12339.70&12.41&12308&12.12&3600.40&4.02&106.70\\
n100q10I&17799&50.46&485.19&16602&40.34&4603.72&&13540.70&14.46&13469&13.86&3600.81&4.15&104.60\\
n100q10J&20044&75.98&459.71&14988&31.59&4578.66&&12491.30&9.67&12462&9.41&3600.54&3.72&96.90\\
\hline
\multicolumn{1}{c}{Avg.}&-&55.84&483.13&-&32.54&4605.65&&-&11.67&-&11.33&3549.54&4.21&101.39\\
\hline
n100q30A&12175&62.41&518.46&8033&7.16&4634.94&&7820.00&4.31&7820&4.31&225.57&1.47&91.00\\
n100q30B&11066&48.70&537.77&10223&37.37&4657.54&&8094.50&8.77&8094&8.76&570.56&0.26&93.50\\
n100q30C&12106&50.14&430.96&9149&13.47&4549.08&&8505.70&5.49&8503&5.46&480.66&0.70&91.80\\
n100q30D&11317&45.52&589.95&9690&24.60&4706.43&&8339.90&7.24&8336&7.19&602.54&0.52&91.10\\
n100q30E&10446&35.42&451.91&8479&9.92&4569.21&&7992.90&3.62&7986&3.53&403.74&1.03&95.90\\
n100q30F&11960&61.14&373.85&8281&11.57&4497.73&&8028.60&8.17&8020&8.05&316.18&1.24&93.40\\
n100q30G&11290&46.31&490.94&8872&14.97&4609.88&&8075.00&4.64&8075&4.64&246.69&0.24&90.00\\
n100q30H&11144&42.19&387.82&8944&14.12&4503.89&&8257.00&5.35&8257&5.35&630.06&0.52&97.00\\
n100q30I&12112&45.32&450.68&9189&10.25&4568.80&&8674.60&4.08&8652&3.81&547.99&1.61&96.50\\
n100q30J&10636&44.70&560.37&9014&22.63&4678.90&&7923.00&7.79&7923&7.79&475.18&0.36&90.40\\
\hline
\multicolumn{1}{c}{Avg.}&-&48.19&479.27&-&16.61&4597.64&&-&5.95&-&5.89&449.92&0.80&93.06\\
\hline
n100q45A&8494&18.12&757.56&8103&12.69&4876.10&&7632.00&6.14&7632&6.14&179.88&0.12&91.00\\
n100q45B&8838&27.33&391.93&8020&15.54&4508.82&&7660.00&10.36&7660&10.36&260.68&0.36&92.50\\
n100q45C&10056&30.21&435.07&8270&7.09&4551.96&&7993.00&3.50&7993&3.50&189.28&0.97&90.40\\
n100q45D&9442&27.15&614.19&8535&14.94&4729.85&&7914.70&6.58&7900&6.38&318.62&0.27&88.20\\
n100q45E&10258&39.47&613.77&7864&6.92&4729.85&&7835.00&6.53&7835&6.53&171.33&7.82&94.10\\
n100q45F&10348&37.36&485.60&7817&3.76&4602.90&&7731.00&2.62&7731&2.62&175.23&1.07&93.30\\
n100q45G&9856&31.59&668.00&8286&10.63&4791.88&&7864.00&4.99&7864&4.99&139.25&0.10&90.00\\
n100q45H&9506&25.94&481.49&7796&3.28&4597.56&&7740.00&2.54&7740&2.54&266.65&5.06&94.00\\
n100q45I&10334&32.23&426.85&8667&10.90&4543.33&&8042.20&2.90&8037&2.84&288.97&0.21&96.10\\
n100q45J&9021&27.53&536.95&7860&11.12&4656.31&&7588.50&7.28&7566&6.96&260.60&0.41&89.70\\
\hline
\multicolumn{1}{c}{Avg.}&-&29.69&541.14&-&9.69&4658.86&&-&5.34&-&5.29&225.05&1.64&91.93\\
\hline
n100q1000A&8447&18.81&682.38&8199&15.32&4803.38&&7453.00&4.83&7453&4.83&132.76&0.10&92.00\\
n100q1000B&8669&22.76&543.52&8183&15.88&4664.11&&7491.00&6.08&7491&6.08&172.13&0.09&93.30\\
n100q1000C&8692&15.81&647.05&8673&15.56&4766.82&&7898.50&5.24&7895&5.19&118.31&0.09&90.10\\
n100q1000D&10116&41.08&489.71&8363&16.63&4605.78&&7572.80&5.61&7565&5.50&163.68&0.11&88.80\\
n100q1000E&8922&17.62&355.78&8071&6.40&4472.26&&7771.00&2.44&7771&2.44&125.79&0.09&94.30\\
n100q1000F&10348&42.47&483.54&8053&10.87&4602.49&&7648.00&5.30&7648&5.30&129.92&0.17&93.10\\
n100q1000G&10954&45.12&432.19&8516&12.82&4549.49&&7817.50&3.57&7813&3.51&96.81&0.06&90.10\\
n100q1000H&9275&22.86&397.68&7786&3.14&3374.94&&7593.00&0.58&7593&0.58&185.62&0.36&94.00\\
n100q1000I&10145&31.58&399.32&8461&9.74&4517.45&&7975.00&3.44&7975&3.44&118.25&0.13&96.00\\
n100q1000J&9178&30.27&446.57&8655&22.84&4565.93&&7315.00&3.82&7315&3.82&117.66&0.06&89.40\\
\hline
\multicolumn{1}{c}{Avg.}&-&28.84&487.77&-&12.92&4492.27&&-&4.09&-&4.07&136.09&0.13&92.11\\
\hline
\end{tabular}}
\label{tab:n100-a1}}
\end{table}

\begin{table}[!ht]
\caption{Detailed results for $n = 100$ and $\alpha = 3$}
\centering
{\small
{\renewcommand{\arraystretch}{1.1}
\renewcommand{\tabcolsep}{0.074cm}
\begin{tabular}{lcccccccccccccc}
\hline\noalign{\smallskip}
& \multicolumn{6}{c}{\textbf{Chemla}} & & & & & &&\\
& \multicolumn{6}{c}{\textbf{et al. 2013b}} &&\multicolumn{7}{c}{\up{\textbf{ILS$_{\text{SBRP}}$}}} \\
\cline{2-7}
\cline{9-15}
\textbf{Instance} &&\textbf{UB1}&\textbf{UB1} &&\textbf{UB2}&\textbf{UB2} & &&\textbf{Avg.} &&\textbf{Best}&\textbf{Avg.}&\textbf{Avg.} \\
& \textbf{UB1}&\textbf{Gap}&\textbf{Time}&\textbf{UB2}&\textbf{Gap}&\textbf{Time} &&\up{\textbf{Avg.}}&\textbf{Gap}&\up{\textbf{Best}}&\textbf{Gap}&\textbf{Time}&\textbf{TT$_{\text{UB2}}$}&\up{\textbf{Avg.}} \\
&&\textbf{(\%)}&\textbf{(s)} &&\textbf{(\%)}&\textbf{(s)} &&\up{\textbf{Sol.}}&\textbf{(\%)}&\up{\textbf{Sol.}}&\textbf{(\%)}&\textbf{(s)}&\textbf{(s)}&\up{\textbf{NV}} \\
\hline
n100q10A&36057&60.84&5133.28&28277&26.14&9301.11&&24121.60&7.60&24014&7.12&3610.42&22.11&97.30\\
n100q10B&47107&71.99&7200.56&35199&28.51&11434.53&&29709.60&8.47&29438&7.48&3631.88&49.10&102.20\\
n100q10C&50606&72.67&486.01&35779&22.08&4682.60&&31802.50&8.51&31540&7.62&3623.25&67.37&100.70\\
n100q10D&47489&51.18&7768.73&37972&20.88&12010.51&&33846.70&7.75&33654&7.14&3641.22&92.56&98.40\\
n100q10E&41002&75.89&776.05&30222&29.65&4938.55&&25092.60&7.64&24917&6.89&3613.53&23.73&105.70\\
n100q10F&43544&86.70&459.30&28488&22.14&4660.42&&25307.10&8.51&25176&7.94&3619.96&50.33&101.30\\
n100q10G&38539&69.26&454.37&28822&26.58&4629.19&&24877.20&9.26&24642&8.23&3616.24&34.89&100.10\\
n100q10H&44411&67.69&587.89&33853&27.83&4816.94&&29039.70&9.65&28794&8.72&3615.54&63.92&106.70\\
n100q10I&48727&65.76&5187.92&37199&26.54&9423.95&&32380.00&10.15&32023&8.94&3643.19&62.52&104.60\\
n100q10J&45590&66.76&355.78&34086&24.68&4543.33&&29373.50&7.44&29147&6.61&3630.50&71.32&96.90\\
\hline
\multicolumn{1}{c}{Avg.}&-&68.87&2840.99&-&25.50&7044.11&&-&8.50&-&7.67&3624.57&53.79&101.39\\
\hline
n100q30A&16110&55.73&441.64&13366&29.20&4560.58&&11278.00&9.02&11258&8.83&3550.25&2.40&91.00\\
n100q30B&18739&68.61&346.74&15537&39.80&4464.04&&12650.70&13.83&12605&13.41&3600.72&4.28&93.50\\
n100q30C&18871&61.46&437.53&16116&37.89&4577.84&&13231.50&13.21&13224&13.14&3600.42&3.55&91.80\\
n100q30D&18262&49.67&722.23&16419&34.56&4865.01&&13786.00&12.98&13783&12.96&3600.72&3.61&91.10\\
n100q30E&16867&69.83&368.10&13118&32.08&4486.23&&10984.10&10.59&10954&10.29&3194.65&3.51&95.90\\
n100q30F&14838&46.49&453.55&13674&35.00&4569.62&&11245.80&11.02&11191&10.48&3600.31&2.21&93.40\\
n100q30G&16772&69.37&381.66&13262&33.92&4505.12&&11183.30&12.93&11160&12.70&3404.20&3.92&90.00\\
n100q30H&15609&40.53&543.52&14495&30.50&4660.00&&12332.80&11.03&12296&10.70&3600.52&4.03&97.00\\
n100q30I&19159&57.85&266.63&16620&36.93&4410.22&&13519.20&11.38&13469&10.97&3600.51&3.48&96.50\\
n100q30J&20044&78.22&459.71&15423&37.13&4578.66&&12513.00&11.26&12462&10.81&3600.43&3.17&90.40\\
\hline
\multicolumn{1}{c}{Avg.}&-&59.78&442.13&-&34.70&4567.73&&-&11.73&-&11.43&3535.27&3.42&93.06\\
\hline
n100q45A&11372&36.43&717.71&10694&28.30&4833.78&&9229.40&10.73&9192&10.28&1597.41&1.78&91.00\\
n100q45B&16114&74.68&409.59&14520&57.40&4526.08&&10233.70&10.93&10209&10.67&2955.64&0.99&92.50\\
n100q45C&14817&50.09&516.00&13243&34.15&4633.30&&10871.50&10.12&10815&9.55&3093.03&1.55&90.40\\
n100q45D&15845&62.58&424.79&15845&62.58&4540.87&&11143.90&14.34&11103&13.92&3562.47&1.03&88.20\\
n100q45E&11628&32.52&602.68&11400&29.92&4718.34&&9521.70&8.51&9498&8.24&1687.03&1.60&94.10\\
n100q45F&12821&50.90&446.16&12243&44.10&4742.17&&9437.70&11.08&9398&10.62&1663.30&1.09&93.30\\
n100q45G&14829&71.60&436.30&10827&25.29&4553.60&&9445.00&9.30&9445&9.30&1215.65&2.09&90.00\\
n100q45H&13072&42.35&652.80&12319&34.15&4768.87&&10226.60&11.37&10206&11.14&2391.88&2.09&94.00\\
n100q45I&14366&46.89&387.00&14366&46.89&4503.07&&10864.10&11.08&10841&10.85&3066.06&1.68&96.10\\
n100q45J&13989&53.75&494.64&11850&30.24&4611.12&&10138.10&11.43&10131&11.35&2469.49&2.09&89.70\\
\hline
\multicolumn{1}{c}{Avg.}&-&52.18&508.77&-&39.30&4643.12&&-&10.89&-&10.59&2370.20&1.60&91.93\\
\hline
n100q1000A&9402&29.43&469.57&8017&10.36&4590.16&&7457.20&2.66&7453&2.60&117.51&0.08&92.00\\
n100q1000B&8793&25.68&506.96&7595&8.55&4627.14&&7491.00&7.07&7491&7.07&182.14&0.82&93.30\\
n100q1000C&9312&21.43&422.74&8554&11.55&4541.69&&7904.30&3.08&7895&2.96&118.19&0.07&90.10\\
n100q1000D&9832&33.73&531.61&7595&3.31&4649.73&&7574.00&3.02&7565&2.90&154.08&26.25&88.80\\
n100q1000E&8922&15.87&355.78&8071&4.82&707.85&&7771.00&0.92&7771&0.92&119.50&0.30&94.30\\
n100q1000F&9371&28.57&721.41&8783&20.50&4838.30&&7648.00&4.93&7648&4.93&124.09&0.06&93.10\\
n100q1000G&10954&46.16&431.78&8219&9.67&4549.08&&7816.50&4.30&7813&4.25&103.84&0.18&90.10\\
n100q1000H&8829&20.19&549.28&8488&15.55&4666.17&&7593.00&3.37&7593&3.37&174.74&0.08&94.00\\
n100q1000I&10664&40.10&306.07&8149&7.06&4421.73&&7975.00&4.78&7975&4.78&124.87&0.38&96.00\\
n100q1000J&8311&20.04&534.49&7976&15.20&4650.56&&7315.00&5.65&7315&5.65&111.68&0.12&89.40\\
\hline
\multicolumn{1}{c}{Avg.}&-&28.12&482.97&-&10.66&4224.24&&-&3.98&-&3.94&133.06&2.83&92.11\\
\hline
\end{tabular}}}
\label{tab:n100-a3}
\end{table}

\section{Concluding Remarks}\label{sec:Conclusions}

In this work we proposed a hybrid ILS algorithm that was especially designed to solve a challenging single-vehicle SBRP variant. Extensive computational experiments were conducted on 980 instances from the literature ranging from 20 to 100 stations. The results were compared with those reported in \citet{chemla} and \citet{erdogan}. For the 900 instances containing up to 60 stations, the proposed heuristic, called ILS$_{\text{SBRP}}$, was capable of finding 796 out of 823 known optimal solutions (97\%)  and improving the result of 61 out of 77 open instances (79\%). Our algorithm only failed to be at least equal to the best known solution in 27 instances (3\%). In addition, the average gap of the average solutions found by ILS$_{\text{SBRP}}$ and the lower bound reported in \cite{erdogan}, for each instance group, was always smaller than 0.7\%, thus ratifying the robustness of our heuristic. As for the 80 instances involving 100 stations, ILS$_{\text{SBRP}}$ outperformed the best heuristics available for the problem by considerably improving the best known solution for all instances.

Future work may include the development of an enhanced procedure to verify whether or not the solution is feasible. Currently, this is the most time consuming part of the algorithm, where we use a relatively costly  max-flow based procedure \citep{chemla} for performing this task. Hence, any improvement on this procedure could possibly lead to an improvement on the CPU time. Also, other type of hybridizations could be experimented by combining, for example, efficient exact algorithms with the heuristic suggested in this work.

\section*{Acknowledgments}

This research was partially supported by the Conselho Nacional de Desenvolvimento Cient{\'i}fico e Tecnol{\'o}gico ́(CNPq), grant 305223/2015-1, and by the Comiss\~ao de Aperfei\c coamento de Pessoal de N\'ivel Superior (CAPES), grant PVE A007$\_$2013.

\clearpage

\bibliographystyle{mmsbib}
\bibliography{paper}









\clearpage

\begin{appendices}

\section{Checking feasibility}\label{app:checking}

Let $L = i_{1}, i_{2}, ..., i_{k}$ be a sequence of vertices, where $i_{1} = i_{k} = 0$. A directed graph can be built using $p_{i}, p^{\prime}_{i}$, and $q_{i}$ for each $i$ in the sequence, as follows:
\begin{itemize}
\item Let $s$ be the source of the flow network, and for each vertex $i$ representing the first occurrence of each station in the sequence let us define a set of arcs $u_{i}$ with capacity $p_{i}$;
\item Let $t$ be the sink of the flow network, and for each $i^{\prime}$ representing the last occurrence of each station in the sequence let us define a set of arcs $w_{i^{\prime}}$ with capacity $p^{\prime}_{i}$;
\item For each $j = 2, ..., k-1$ let us define an arc $b_{j,j+1}$ with capacity $Q$; and
\item If a station $i$ is visited more than once, let us define an arc $d_{e,e+1}$ with capacity $q_{i}$, between the $e{th}$ and $(e+1){th}$ visits to $i$.
\end{itemize}

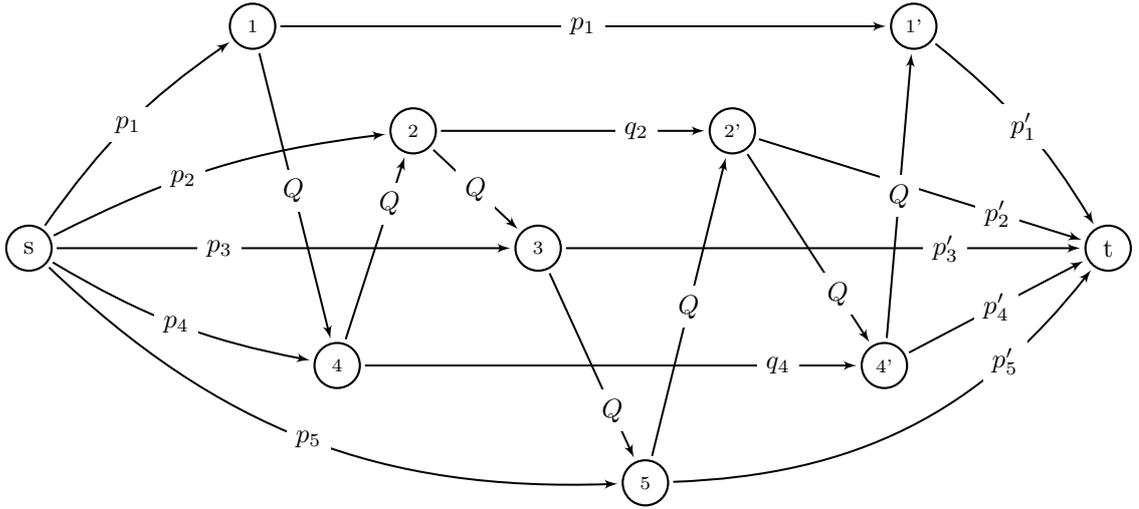
\begin{figure}[!ht]
	\centering
	\begin{tikzpicture}[>=latex',node distance=2.2cm,style={
main node/.style={circle,draw,minimum size=17,inner sep=1pt,thick}}]
\SetUpEdge[lw = 0.75pt,color = black,labelcolor = white]
\tikzset{EdgeStyle/.style={->,shorten <=1.5pt,shorten >=1.5pt}}
\scriptsize
	\node[main node] 				(0) {\normalsize{s}};
	\node[main node, right = 1cm] 	(1) [above right=2.5cm and 2.5cm of 0] {1};
	\node[main node, right = 2.5cm]   (4) [below right of=0] {4};
	\node[main node, right = 2.5cm]	(5) [below right of=4] {5};
	\node[main node, right = 4.5cm] 	(3) [right of=0] {3};
	\node[main node, right = 3.5cm] 	(2) [above right of=0] {2};
    \node[main node, right = 2cm] 	(22) [right of=2] {2'};
    \node[main node, right = 6.5cm] 	(11) [right of=1] {1'};
    \node[main node, right = 5cm] 	(44) [right of = 4] {4'};
    \node[main node, right = 12cm] 	(999) [right of=0] {\normalsize{t}};
\normalsize
    \Edge[style={bend left=10},label={$p_1$}](0)(1)
    \Edge[style={bend left=10,pos=0.4},label={$p_2$}](0)(2)
    \Edge[style={bend left=0,pos=0.35},label={$p_3$}](0)(3)
    \Edge[style={bend right=10},label={$p_4$}](0)(4)
    \Edge[style={bend right=23},label={$p_5$}](0)(5)
        
    \Edge[style={bend right=0,pos=0.48},label={$Q$}](1)(4)
    \Edge[style={bend right=0},label={$p_1$}](1)(11)   
    
    \Edge[style={bend right=0,near end},label={$q_2$}](2)(22)   
    \Edge[style={bend right=0},label={$Q$}](2)(3)   
    
    \Edge[style={bend right=0,near end},label={$Q$}](3)(5)   
    \Edge[style={bend right=0,near end},label={$p^{\prime}_3$}](3)(999)           
    
    \Edge[style={bend right=0,near end},label={$Q$}](4)(2)
    \Edge[style={bend right=0,pos=0.85},label={$q_{4}$}](4)(44)       

    \Edge[style={bend right=0,near end},label={$Q$}](22)(44)
    \Edge[style={bend right=0,near end},label={$p^{\prime}_2$}](22)(999)  

    \Edge[style={bend right=0},label={$Q$}](44)(11)
    \Edge[style={bend right=0},label={$p^{\prime}_4$}](44)(999)
    
    \Edge[style={bend right=0,bend left = 10},label={$p^{\prime}_1$}](11)(999)
    
    \Edge[style={bend right=0},label={$Q$}](5)(22)    
    \Edge[style={bend right=0,bend right = 25, near end},label={$p^{\prime}_5$}](5)(999);

\end{tikzpicture}
	\caption{Flow network for feasibility checking}
	\label{figure:checker}
\end{figure}

By computing an $s$--$t$ maximum flow, one can find optimal bike displacements along the sequence $L$. For each station $i$, let us define $\hat{p_{i}}$ and $\hat{p}^{\prime}_{i}$, respectively, as the resulting $s$--$t$ flow on arcs $u_i$ and $w_i$. 
Flow on arcs $b_{j,j+1}$ indicates the number of bikes from $i_{j}$ to $i_{j+1}$ and flow on arcs $d_{e,e+1}$ denotes the quantity of bikes remaining in a station $i$ after the $e{th}$ visit and before the $(e+1){th}$ visit. Figure \ref{figure:checker} depicts a flow network for a sequence $L = 0,1,4,2,3,5,2,4,1,0$, where $s$ and $t$ correspond to the depot.

\citet{chemla} states that sequence $L$ induces a feasible solution when $\hat{p_{i}} = p_{i}$, for each station $i$ in the sequence. Also, if a vertex $i^{\prime}$ is not in $L$, then $\hat{p}_{i^{\prime}} = \hat{p}^{\prime}_{i^{\prime}} = 0$.

\section{Detailed results}
\label{app:results}

This appendix presents the detailed results found by ILS$_\text{SBRB}$, as well as those obtained by the exact algorithm of \cite{erdogan}, for the instances with up to 60 stations.


\renewcommand{\tabcolsep}{0.214cm}
\singlespacing

{\footnotesize
\begin{center}
\begin{longtable}{lcccccccccccc}
\caption{Detailed results for $n \in \{20, 30, 40, 50, 60\}$ and $\alpha = 1$} \label{tab:appendix-a1} \\

\hline 
& \multicolumn{4}{c}{\textbf{Erdo\v{g}an}} \\
& \multicolumn{4}{c}{\textbf{et al. 2015}} &&\multicolumn{7}{c}{\up{ILS$_{\textbf{SBRP}}$}} \\
\cline{2-5} \cline{7-13}
\textbf{Instance} & &&\textbf{UB}&\textbf{UB} & &&\textbf{Avg.} &&\textbf{Best}&\textbf{Avg.}&\textbf{}&\textbf{Best} \\
& \textbf{LB}&\textbf{UB}&\textbf{Gap}&\textbf{Time} &&\up{\textbf{Avg.}}&\textbf{Gap.}&\textbf{Best}&\textbf{Gap}&\textbf{Time}&\up{\textbf{Avg.}}&\textbf{of all} \\
& &&\textbf{(\%)} & \textbf{(s)} & & \up{\textbf{Sol.}} & \textbf{(\%)} & & \textbf{(\%)} & \textbf{(s)} & \up{\textbf{NV}} & \textbf{exp.} \\  
\hline
\endfirsthead

\hline 
& \multicolumn{4}{c}{\textbf{Erdo\v{g}an}} \\
& \multicolumn{4}{c}{\textbf{et al. 2015}} & & \multicolumn{7}{c}{\up{ILS$_{\textbf{SBRP}}$}} \\
\cline{2-5} \cline{7-13}
\textbf{Instance} & & & \textbf{UB} & \textbf{UB} & & & \textbf{Avg.} & & \textbf{Best} & \textbf{Avg.} & \textbf{} & \textbf{Best} \\
& \textbf{LB} & \textbf{UB} & \textbf{Gap} & \textbf{Time} & & \up{\textbf{Avg.}} & \textbf{Gap.} & \textbf{Best} & \textbf{Gap} & \textbf{Time} & \up{\textbf{Avg.}} & \textbf{of all} \\
& & & \textbf{(\%)} & \textbf{(s)} & & \up{\textbf{Sol.}} & \textbf{(\%)} & & \textbf{(\%)} & \textbf{(s)} & \up{\textbf{NV}} & \textbf{exp.} \\ 
\hline
\endhead

\hline \multicolumn{13}{r}{{\textit{Continued on next page}}}
\endfoot

\hline
\endlastfoot

n20q10A&4702&4702&0.00&0.26&&4702.00&0.00&4702&0.00&2.93&18.00&4702\\
n20q10B&4769&4769&0.00&0.11&&4769.00&0.00&4769&0.00&4.04&19.00&4769\\
n20q10C&6012&6012&0.00&0.68&&6012.00&0.00&6012&0.00&18.64&24.00&6012\\
n20q10D&5989&5989&0.00&0.20&&5989.00&0.00&5989&0.00&7.77&20.00&5989\\
n20q10E&6245&6245&0.00&1.31&&6245.00&0.00&6245&0.00&8.81&20.00&6245\\
n20q10F&4717&4717&0.00&0.25&&4717.00&0.00&4717&0.00&3.49&22.00&4717\\
n20q10G&5070&5070&0.00&0.07&&5070.00&0.00&5070&0.00&2.74&19.10&5070\\
n20q10H&5542&5542&0.00&0.30&&5575.20&0.60&5542&0.00&2.77&19.90&5542\\
\end{longtable}
\end{center}}

\renewcommand{\tabcolsep}{0.195cm}

{\footnotesize
\begin{center}
\begin{longtable}{lcccccccccccc}
\caption{Detailed results for $n \in \{20, 30, 40, 50, 60\}$ and $\alpha = 3$} \label{tab:appendix-a3} \\

\hline 
& \multicolumn{4}{c}{\textbf{Erdo\v{g}an}} \\
& \multicolumn{4}{c}{\textbf{et al. 2015}} & & \multicolumn{7}{c}{\up{ILS$_{\textbf{SBRP}}$}} \\
\cline{2-5} \cline{7-13}
\textbf{Instance} & & & \textbf{UB} & \textbf{UB} & & & \textbf{Avg.} & & \textbf{Best} & \textbf{Avg.} & \textbf{} & \textbf{Best} \\
& \textbf{LB} & \textbf{UB} & \textbf{Gap} & \textbf{Time} & & \up{\textbf{Avg.}} & \textbf{Gap.} & \textbf{Best} & \textbf{Gap} & \textbf{Time} & \up{\textbf{Avg.}} & \textbf{of all} \\
& & & \textbf{(\%)} & \textbf{(s)} & & \up{\textbf{Sol.}} & \textbf{(\%)} & & \textbf{(\%)} & \textbf{(s)} & \up{\textbf{NV}} & \textbf{exp.} \\ 
\hline
\endfirsthead

\hline 
& \multicolumn{4}{c}{\textbf{Erdo\v{g}an}} \\
& \multicolumn{4}{c}{\textbf{et al. 2015}} & & \multicolumn{7}{c}{\up{ILS$_{\textbf{SBRP}}$}} \\
\cline{2-5} \cline{7-13}
\textbf{Instance} & & & \textbf{UB} & \textbf{UB} & & & \textbf{Avg.} & & \textbf{Best} & \textbf{Avg.} & \textbf{} & \textbf{Best} \\
& \textbf{LB} & \textbf{UB} & \textbf{Gap} & \textbf{Time} & & \up{\textbf{Avg.}} & \textbf{Gap.} & \textbf{Best} & \textbf{Gap} & \textbf{Time} & \up{\textbf{Avg.}} & \textbf{of all} \\
& & & \textbf{(\%)} & \textbf{(s)} & & \up{\textbf{Sol.}} & \textbf{(\%)} & & \textbf{(\%)} & \textbf{(s)} & \up{\textbf{NV}} & \textbf{exp.} \\ 
\hline
\endhead

\hline \multicolumn{13}{r}{{\textit{Continued on next page}}}
\endfoot

\hline
\endlastfoot

n20q10A&9084&9084&0.00&0.13&&9084.00&0.00&9084&0.00&116.11&35.00&9084\\
n20q10B&9883&9883&0.00&0.14&&9883.00&0.00&9883&0.00&129.78&34.30&9883\\
n20q10C&14039&14039&0.00&1.62&&14039.00&0.00&14039&0.00&384.84&42.40&14039\\
n20q10D&14925&14925&0.00&0.15&&14925.00&0.00&14925&0.00&501.00&46.00&14925\\
n20q10E&14379&14379&0.00&0.82&&14379.00&0.00&14379&0.00&319.86&43.00&14379\\
n20q10F&9752&9752&0.00&0.62&&9752.00&0.00&9752&0.00&125.42&36.00&9752\\
n20q10G&10600&10600&0.00&0.19&&10600.00&0.00&10600&0.00&116.54&37.00&10600\\
n20q10H&11895&11895&0.00&0.11&&11895.00&0.00&11895&0.00&228.47&42.00&11895\\
\end{longtable}
\end{center}}
\vspace{-30pt}\noindent\footnotesize{$^*$Instance in which the cost of our best feasible solution is 1 unit smaller than the cost of the optimal solution reported in \cite{erdogan} (possibly due to rounding issues).}
\end{appendices}

\end{document}